\documentclass[twoside,11pt]{article}

\usepackage[abbrvbib,preprint]{jmlr2e}

\usepackage{bm}
\usepackage{wrapfig}
\usepackage{amsmath}
\usepackage{mathtools}
\usepackage{array}

\usepackage{algorithm}
\usepackage{algorithmic}
\usepackage{lipsum}
\usepackage{soul}

\usepackage{xcolor}         
\usepackage{subcaption}
\usepackage{booktabs, makecell}
\usepackage{multirow}

\newcommand{\R}{\mathbb{R}}

\newcommand{\imgline}{\includegraphics[height = 0.12\textwidth]{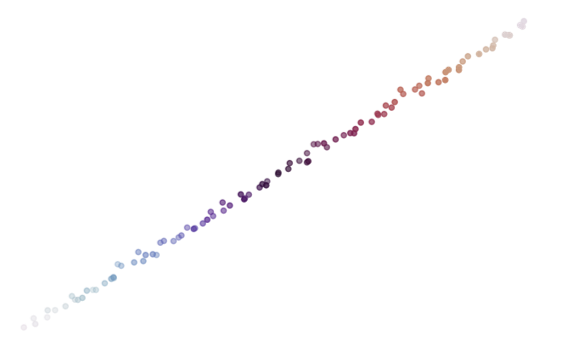}} 
\newcommand{\imgcircle}{\includegraphics[height = 0.12\textwidth]{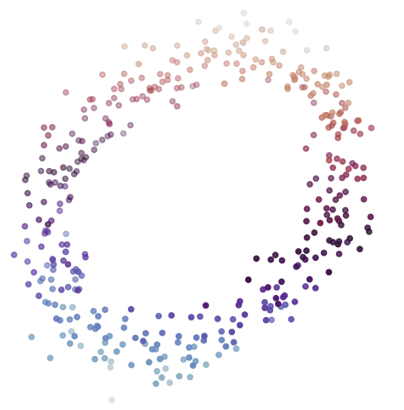}} 
\newcommand{\imgcircletwo}{\includegraphics[height = 0.12\textwidth]{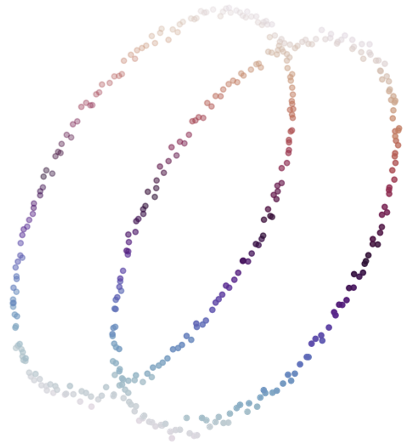}} 
\newcommand{\imgcirclethree}{\includegraphics[height = 0.12\textwidth]{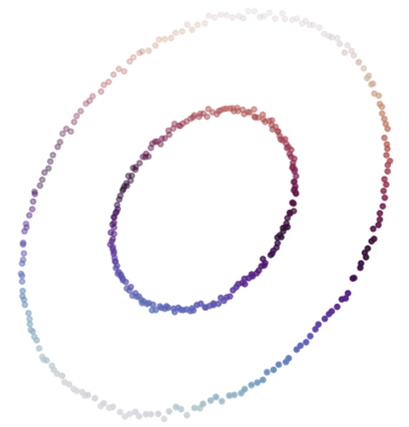}} 
\newcommand{\imgspiral}{\includegraphics[height = 0.12\textwidth]{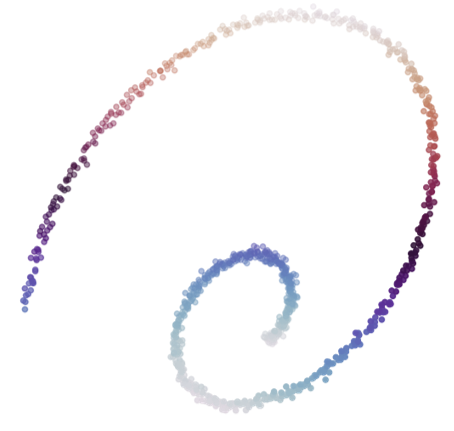}}

\newcommand{\imgHzline}{\includegraphics[height = 0.23\textwidth]{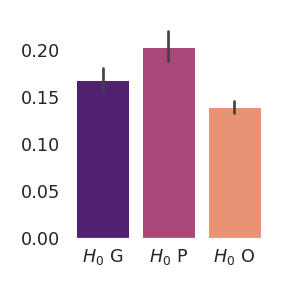}}

\newcommand{\imgHzCircle}{\includegraphics[height = 0.23\textwidth]{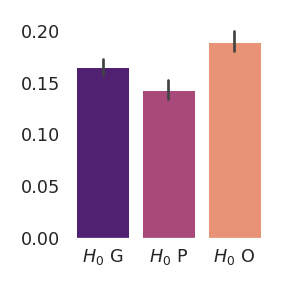}} 

\newcommand{\imgHOneCircle}{\includegraphics[height = 0.23\textwidth]{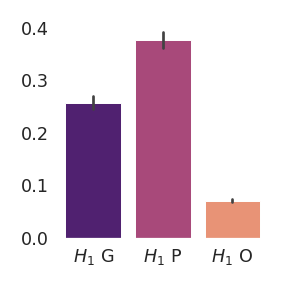}} 

\newcommand{\imgHzCircleii}{\includegraphics[height = 0.23\textwidth]{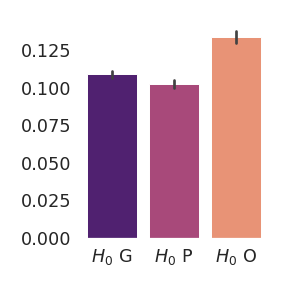}} 

\newcommand{\imgHOneCircleii}{\includegraphics[height = 0.23\textwidth]{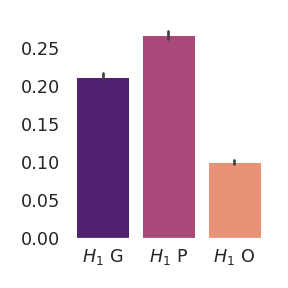}} 

\newcommand{\imgHzCircleiii}{\includegraphics[height = 0.23\textwidth]{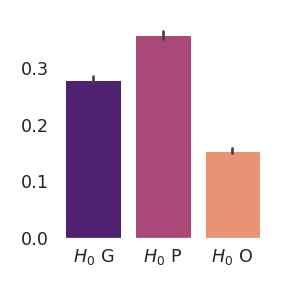}} 

\newcommand{\imgHOneCircleiii}{\includegraphics[height = 0.23\textwidth]{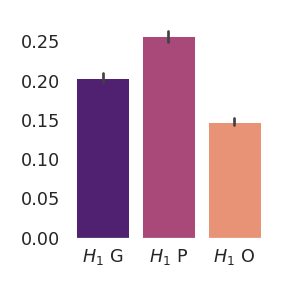}} 

\newcommand{\imgHzSpiral}{\includegraphics[height = 0.23\textwidth]{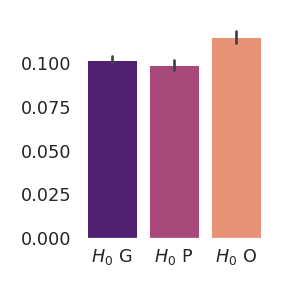}} 

\newcommand{\imgHOneSpiral}{\includegraphics[height = 0.23\textwidth]{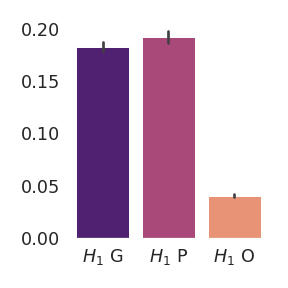}}

\DeclareMathOperator*{\argmin}{arg\,min}

\jmlrheading{1}{2000}{1-48}{4/00}{10/00}{meila00a}{Marina Meil\u{a} and Michael I. Jordan}

\ShortHeadings{EMaP: Explainable AI with Manifold-based Perturbations}{Vu, Mai and Thai}
\firstpageno{1}

\begin{document}

\title{EMaP: Explainable AI with Manifold-based Perturbations}

\author{\name Minh Nhat Vu
        \email minhvu@ufl.edu \\
       \addr Department of Computer and Information Science and Engineering\\
       University of Florida\\
       Gainesville, FL 32611, USA
       \AND
       \name Huy Quang Mai
        \email huyqmai1@gmail.com  \\
       \addr
       Hanoi, 10000, Vietnam
       \AND
       \name My T. Thai
        \email mythai@cise.ufl.edu \\
       \addr Department of Computer and Information Science and Engineering\\
       University of Florida\\
       Gainesville, FL 32611, USA}

\editor{Kevin Murphy and Bernhard Sch{\"o}lkopf}

\maketitle

\begin{abstract}
In the last few years, many explanation methods based on the perturbations of input data have been introduced to improve our understanding of decisions made by black-box models. The goal of this work is to introduce a novel perturbation scheme so that more faithful and robust explanations can be obtained. Our study focuses on the impact of perturbing directions on the data topology. We show that perturbing along the orthogonal directions of the input manifold better preserves the data topology, both in the worst-case analysis of the discrete Gromov-Hausdorff distance and in the average-case analysis via persistent homology. From those results, we introduce EMaP algorithm, realizing the orthogonal perturbation scheme. Our experiments show that EMaP not only improves the explainers' performance but also helps them overcome a recently-developed attack against perturbation-based methods.
\end{abstract}


\section{Introduction}

In recent years, many attempts to explain decisions of deep learning models have been conducted, which resulted in various explanation methods called \textit{explainers}~\citep{Lipton2016IML,Murdoch22071}. A common
technique used by many explainers~\citep{trumbelj2013ExplainingPM,Marco2016,Scott2017} is first to generate some perturbations in the input's space, then forward them through the model and later provide an explanation based on the captured outputs. For that reason, these methods are also known as \textit{perturbation-based explainers}.

Even though the perturbation-generating step has a strong influence on the performance of explainers~\citep{Marco2016,Scott2017}, very few works closely examined this step. Current perturbation schemes often ignore the data topology and distort it significantly as a result. These distortions can considerably degrade explainers' performance since models are not trained to operate on the deformed topology. Additionally, the difference between the perturbations and the original data creates opportunities for malicious intents. For example, the work~\citep{FoolingLIMESHAP} demonstrates that a discriminator trained to recognize the explainer's perturbations can be exploited to fool the explainer.

  \begin{figure}[ht]
     \centering
         \includegraphics[width=0.96\linewidth]{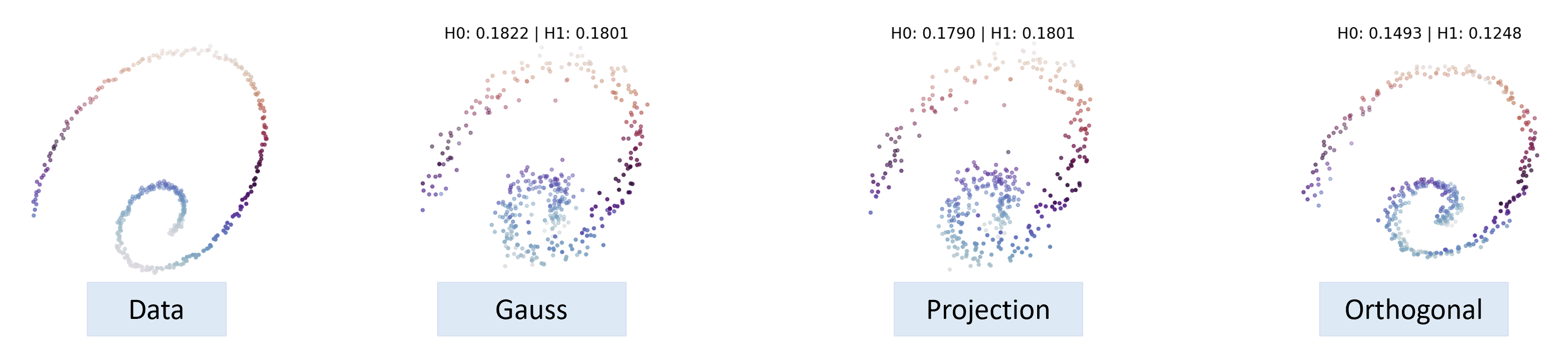}
     \caption{Visualization of perturbations with the same magnitude generated from a point cloud of a 2-dimensional spiral. Perturbations along the orthogonal directions of the data subspace (far-right) result in lower topological distortion, i.e. smaller Bottleneck distances $H_0$ and $H_1$.}
     \label{fig:circles2}
 \end{figure}

 Motivated by that lack of study, our work aims to re-design the perturbation step in an explanation process so that the topological structure of the original data is better preserved. Our key result is that, assuming the input data is embedded in an affine subspace whose dimension is significantly smaller than that of the data dimension, eliminating the perturbations’ components along that affine subspace would better preserve the topological integrity of the original manifold. An illustration of that result is provided in Fig.~\ref{fig:circles2}, which shows that perturbation along the orthogonal directions (i.e. no subspace's directions) results in smaller distortion in the topological structure of the original data, which is reflected in the smaller Bottleneck distances in dimension 0 and 1, denoted by $H_0$ and $H_1$.
 
\begin{figure}[h]
     \centering
         \includegraphics[width=0.99\linewidth]{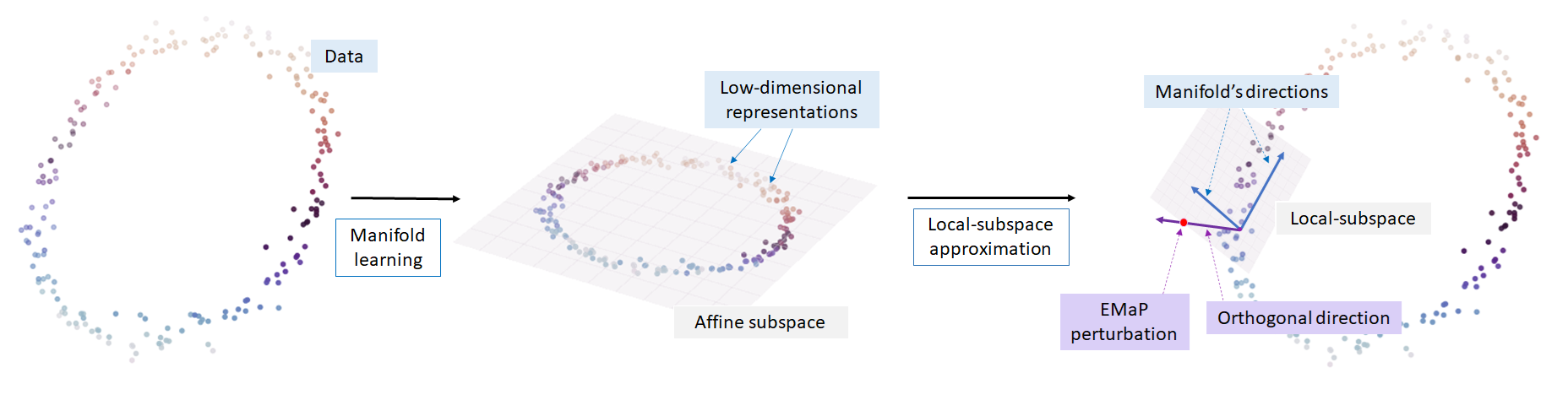}
     \caption{EMaP's perturbation: Assume the data is embedded in a low-dimensional affine subspace (middle figure), EMaP approximates that subspace locally at some given data points and performs perturbation along orthogonal directions of that subspace (right figure).}
     \label{fig:illustration}
 \end{figure}

 Based on that result, we further propose a novel manifold-based perturbation method aiming to preserve the topological structure of the original data, called EMaP. The high-level operations of EMaP are shown in Fig.~\ref{fig:illustration}. Given some sampled data, EMaP first learns a function mapping the samples to their low-dimensional representations in the data subspace. Then, that function is used to approximate a local affine subspace, shortened to local-subspace, containing the data in the neighborhood of the input to be explained. Finally, the EMaP perturbations are generated by adding the noise vectors that are orthogonal to that local-subspace to the data.
 
{\bf Contributions.} \textbf{(a)} We theoretically show that the worst-case discrete Gromov-Hausdorff distance between the data and the perturbations along the manifold's directions is larger than that along the orthogonal directions. \textbf{(b)} The worst-case analysis suggests that eliminating perturbation's components along the manifold's directions can generally better maintain the  topological integrity of the original manifold, i.e. the average-case. We then provide synthetic and real-world experiments based on persistent homology and Bottleneck distance to support that claim. \textbf{(c)} We propose EMaP, an algorithm generating perturbations along the manifold's orthogonal directions for explainers. EMaP first approximates the input's manifold locally at some given data points, called pivots, and the explained data point. The perturbations are then generated along the orthogonal directions of these local-subspaces. EMaP also computes the low-dimensional distances from the perturbations to the explained data point so that the explainers can better examine the model. \textbf{(d)} Finally, we provide experiments on four text datasets, two tabular datasets, and two image datasets, showing that EMaP can improve the explainer's performance and protect explainers from adversarial discriminators.

\paragraph{Organization.} The remainder of the paper is structured as follows. Sections~\ref{sect:related} and~\ref{sect:prelim} briefly discuss related work and preliminaries. Section~\ref{sect:theory} presents our analysis of the discrete Gromov-Hausdorff distances of different perturbation directions, which suggests orthogonal directions are preferable. We strengthen that result with a persistent homology analysis in Section~\ref{sect:bd}. Sections~\ref{sect:Emap} and 
\ref{sect:exp} describe our proposed EMaP algorithm and its experimental results. Section~\ref{sect:conclusion} concludes the paper.

\section{Related work}\label{sect:related}
This work intersects several emerging research fields, including explainers and their attack/defense techniques. Our approach also uses recent results in topological data analysis. We provide an overview of those related work below.

{\bf Perturbation-based explanation methods.} Perturbation-based explainers are becoming more popular among explanation methods for black-box models since they hardly require the any knowledge on the explained model. Notable ones are LIME~\citep{Marco2016}, SHAP~\citep{Scott2017}, and some others~\citep{trumbelj2013ExplainingPM, Zeiler2014, Mukund2017, chang2018explaining, schwab2019cxplain, Lundberg2020}. While they share the same goal to explain the model's predictions, they are not only different in the objectives but also in their perturbation schemes: some zero out features~\citep{Zeiler2014,schwab2019cxplain} or replace features with neutral values~\citep{Marco2016,Mukund2017}, others marginalize over some distributions on the dataset~\citep{Marco2016,Scott2017,Lundberg2020}. There also exist methods relying on separate models to generate perturbations~\citep{trumbelj2013ExplainingPM,chang2018explaining}. The work~\citep{Covert2021} provides a comprehensive survey on those  perturbation-based explanation methods and how they perturb the data.





\textbf{Adversarial attack on explainers.}
We focus on the attack framework~\citep{FoolingLIMESHAP}, in which the adversary intentionally hides a biased model from the explainer by training a discriminator recognizing its query. The framework will be discussed in details in Section.~\ref{sect:prelim}. There are other emerging attacks on explainers focusing on modifying the model's weights and tampering with the input data~\citep{Ghorbani_Abid_Zou_2019, Dombrowski19, Heo2019FoolingNN, Dimanov2020YouST}.

\textbf{Defense techniques for perturbation-based explainers.} Since most attacks on perturbation-based explainers were only developed recently, defense techniques against them are quite limited. Existing defenses generate perturbations either from carefully sampling the training data~\citep{Joymallya} or from learning some generative models~\citep{Saito2020ImprovingLR,Domen}. The advantage of EMaP is that it does not require any generative model, which not only reduces the attack surface but also allows theoretical study on the perturbations.

\textbf{Topological Data Analysis.}
Topological data analysis (TDA) is an emerging field in mathematics, applying the techniques of topology (which was traditionally very theoretical) to real-world problems. Notable applications are data science, robotics, and neuroscience. TDA uses deep and powerful mathematical tools in algebraic topology to explore topological structures in data and to provide insights that normal metric-based methods fail to discern. The most common tool in the TDA arsenal is persistent homology, developed in the early 2000s by Gunnar Carlsson and his collaborators. We refer readers to \citep{ghrist2014elementary,edelsbrunner2010computational} for an overview of both persistent homology and TDA as a whole.

\section{Preliminaries}\label{sect:prelim}

We use the standard setting of the learning tasks where the set of input $X$ is sampled from a distribution on $\mathbb{R}^N$. $X$ is also assumed to be in a manifold embedded in an affine subspace $\R^V$, where $V$ is much smaller than $N$. We also consider a black-box classifier $f$ mapping each input $x\in X$ to a prediction $y$, a local explainer $g$, a (adversarial) discriminator $\mathcal{D}$, and a masking-model $f'$. 


\textbf{Explainers.}
An explanation of prediction $y = f(x)$ can be obtained by running an explainer $g$ on $x$ and $f$. We denote such explanation by $g(x)$. In additive feature attribution methods, the range of $g(x)$ is a set of features' importance scores. We focus our analysis in the class of perturbation-based explanation, i.e. the importance scores are computed based on the model's predictions of some perturbations of the input data. The perturbations are commonly generated by perturbing some samples in $X$. We denote the perturbations by $X_r$, where $r \geq 0$ specifies the amount of perturbation. A more rigorous definition for this notation will be provided in Section~\ref{sect:theory}.

\begin{figure}{h}
     \centering
         \includegraphics[width=0.7\linewidth]{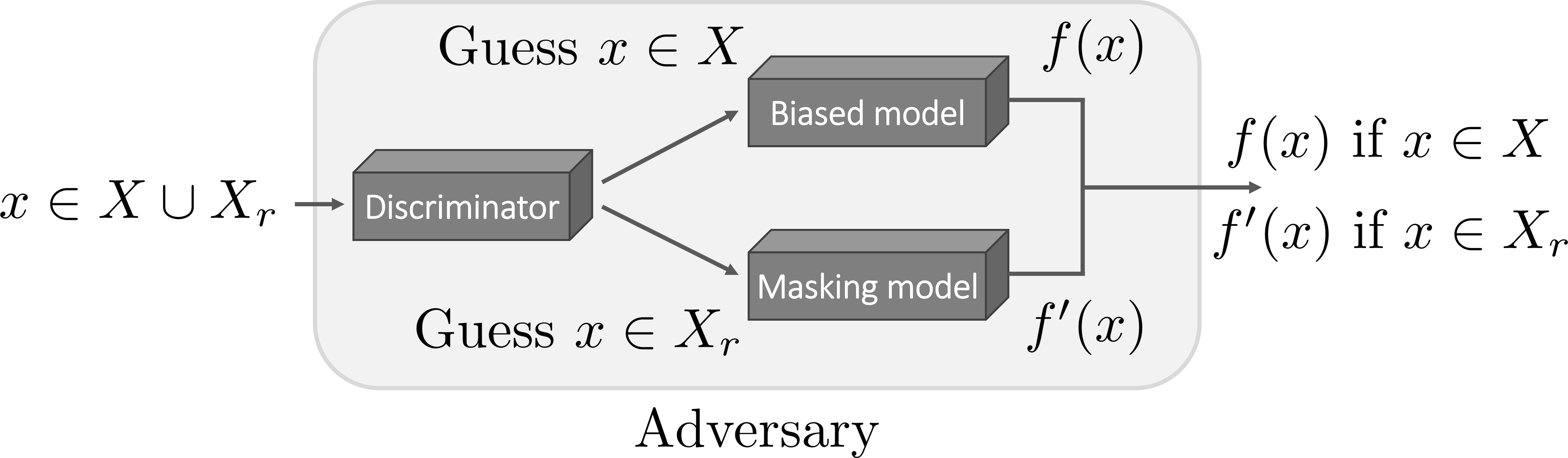}
     \caption{The discriminator-based attack framework: By recognizing and forwarding the perturbations $X_r$ generated by an explainer to the masking model $f'$, the biased-model $f$ can be deployed without detection.}
     \label{fig:attack}
 \end{figure}

Our experiments are mainly on the LIME explainer~\citep{Marco2016} because of its nice formulation, popularity, and flexibility. The output of LIME is typically a linear model $h$ whose coefficients are the importance score of the features:
\begin{align*}
    \argmin_{h\in \mathcal{H}} \sum_{x \in X_r}\mathcal{L}(f(x'),h(x'),\kappa(x')) + \Omega(h), 
\end{align*}
where $\mathcal{H}$ is the searching space, $\kappa$ is the weight function measuring the similarity between $x'$ and the explained input $x$, $\mathcal{L}$ is the loss function measuring the difference between 
$f(x')$ and the linear approximation $h(x')$, and $\Omega(h)$ is a function measuring the complexity of $h$.

\textbf{Attack framework.} We study the discriminator-based attack framework introduced by~\citep{FoolingLIMESHAP}, which is illustrated in Fig.~\ref{fig:attack}. In the framework, there is an adversary with an incentive to deploy a biased-model $f$. This adversary can bypass detection of the explainer by forwarding the explainer's perturbations in $X_r$ to a masking model $f'$. The decision whether to forward the inputs to the masking model is made by a discriminator $\mathcal{D}$. Thus, the success of the attack is determined by the capability to distinguish $X_r$ from $X$ of the discriminator $\mathcal{D}$. Intuitively, if the explainer can craft an $X_r$ similar to $X$, it not only improves the explainer's performance but also prevents the adversary from hiding its bias.

\section{Analysis of Discrete Gromov-Hausdorff distances of perturbations}\label{sect:theory}

We consider the following perturbation problem: Given a manifold embedded in $\R^N$, how do we perturb it so that we preserve as much topological information as possible? More concretely, given a finite set of points sampled from such a manifold, is there a consistent method to perturb the original dataset while preserving some notion of topology? 

To begin talking about differences between (metric) spaces, we need to introduce a notion of distance between them. One such commonly used distance is the \textbf{Gromov-Hausdorff distance}. Intuitively, a small Gromov-Hausdorff distance means that the two spaces are very similar as metric spaces. Thus, we can focus our study on the Gromov-Hausdorff distances between the data and different perturbation schemes. However, as it is infeasible to compute the distance in practice, we instead study an approximation of it, which is the discrete Gromov-Hausdorff distance. Specifically, we show that, when the perturbation is significantly small, the worst-case discrete Gromov-Hausdorff distance resulted from orthogonal perturbation is smaller than that of projection perturbation, i.e. perturbation along the manifold (Theorem~\ref{theorem:proj_vs_orth}). The proof of that claim relies on Lemma~\ref{lemma:first}, which states that, with a small perturbation, the discrete Gromov-Hausdorff distance between the original point cloud and the perturbation point cloud equals to the largest change in the distances of any pair of points in the original point cloud. With the Lemma, the problem of comparing point clouds is further reduced to the problem of comparing the change in distances.

We now state the formal definitions. Let $(M,d)$ be a metric space. For a subset $S \subseteq M$ and a point $y \in M$, the distance between $S$ and $y$ is given by $d(S,y) := \inf_{x\in S} d(x,y)$.

\begin{definition}[Hausdorff distance]\citep{tuzhilin2016invented} Let $S$ and $S'$ be two non-empty subsets of a metric space $(M,d)$. The Hausdorff distance between $S$ and $S'$, denoted by $d_H(S,S')$ is:
\begin{align*}
    d_H (S,S') := \max \big( \sup_{x \in S} d(S', x), \sup_{y \in S'} d(S,y) \big)
\end{align*}

\end{definition}

\begin{definition} [Gromov-Hausdorff distance]\citep{tuzhilin2016invented}
Let $X,Y$ be two compact metric spaces. The Gromov-Hausdorff distance between $X$ and $Y$ is given by: 
\begin{align*}
    d_{GH}(X,Y) = \inf_{f,g} d_H(f(X),f(Y)) 
\end{align*}
where the infimum is taken over all metric spaces $M$ and all isometric embeddings $f: X \to M$, $g: Y \to M$.
\end{definition}

Even though the Gromov-Hausdorff distance is mathematically desirable, it is practically non-computable since the above infimum is taken over all possible metric spaces. In particular, this includes the computation of the Gromov-Hausdorff distance between any two point clouds. In 2004, Memoli and Sapiro ~\citep{memoli:compare} addressed this problem by using a discrete approximation of Gromov-Hausdorff, which looks at the distortion of pairwise distances over all possible matchings between the two point clouds. Formally, given two finite sets of points $X = \{x_1,..., x_n\}$ and $Y = \{y_1,..., y_n\}$ in a metric space $(M,d)$, the discrete Gromov-Hausdorff distance between $X$ and $Y$ is given by
\begin{equation}
    d_J(X,Y) = \min_{\pi \in S_n} \max_{i,j} \frac{1}{2} \big | d(x_i,x_j) - d(y_{\pi(i)},y_{\pi(j)}) \big | \label{eq:dJ}
\end{equation}
where $S_n$ is the set of all $n$-permutations. 

Let $X = \{x_1,..., x_k\} \in \R^V \subseteq \R^N$ be a point cloud contained in some affine subspace $\R^V$ of $\R^N$. We say $X$ is generic if the pairwise distances between the points in $X$ are not all equal, i.e. there exist some points $x_{i_1},x_{i_2},x_{i_3},x_{i_4} \in X$ such that $d(x_{i_1},x_{i_2}) \neq d(x_{i_3},x_{i_4}) $.

Let $X_r$ be a finite set of points in $\R^N$ s.t. for every $x_i \in X$, there exists a unique $\tilde{x}_i \in X_r$ such that $d(x_i,\tilde{x}_i) = r $. $X_r$ realizes a perturbation of $X$ with the radius of perturbation being equal to $r$. We also denote $X^\perp_r$ (resp. $X^\text{Proj}_r$) as a finite set of points in $\R^N$ such that for every $x_i \in X$, there exists a unique $\tilde{x}_i \in X^\perp_r$ such that $d(x_i,\tilde{x}_i) = r $ and  $\overline{x_i \tilde{x}_i} \perp \R^V$ (resp. $\overline{x_i \tilde{x}_i} \subset \R^V$), where $\overline{x_i \tilde{x}_i}$ denotes the line connecting the points $x_i$ and $\tilde{x}_i$. We are now ready to state the following key theorem: 

\begin{theorem}\label{theorem:proj_vs_orth}
Given a generic point-cloud $X = \{x_1,..., x_k\} \in \R^V \subset \R^N$, there exists an $r_0 > 0$ such that for any $r < r_0$ and for any instances of $X_r^\perp$, there exists an $X_r^\textup{Proj}$ such that:
\begin{equation*}
    d_J (X, X_r^\perp ) \leq d_J (X, X_r^\textup{Proj})
\end{equation*}
\end{theorem}

We prove Theorem \ref{theorem:proj_vs_orth} by showing the following lemma:

\begin{lemma} \label{lemma:first}
There exists an $\epsilon > 0$ such that for any $r < \epsilon $, we have
\begin{equation*}
d_J (X, X_r) = \frac{1}{2} \max_{i,j}{|d(x_i,x_j)- d(\tilde{x}_i,\tilde{x}_j)|}
\end{equation*}
for any $X_r$.
\end{lemma}

To prove Lemma~\ref{lemma:first}, we show that, for a small enough $\epsilon$, the optimal permutation in Eq. (\ref{eq:dJ}) is the identity $\pi(i) = i$. Thus, the minimization in the computation of $d_{J}$ can be eliminated. The detail is shown in the following.

\begin{proof} \textbf{of Lemma~\ref{lemma:first}.}
Given a permutation $\pi \in S_n$ and two point clouds $X,Y$ of the same cardinality, denote: 
\[
D_\pi (X,X_r) = \frac{1}{2} \max_{i,j} |d(x_i,x_j) - d(\tilde{x}_{\pi(i)}, \tilde{x}_{\pi(j)})| 
\]
Let $N_0$ be the set of permutations $\pi \in S_n$ such that $D_\pi(X,X) = 0$. Let $N_1 = S_n \backslash N_0$.
Since $X$ is generic, $N_0$ does not include all of $S_n$ and $N_1 \neq \emptyset$.

Let $\delta = \min_{\pi \in N_1} D_\pi(X,X) > 0$. We claim that choosing $\epsilon = \frac{\delta}{4}$ proves the lemma. To be more precise, the radius of perturbation $r$ is chosen such that $r <\epsilon = \frac{\delta}{4}$, i.e. $4r < \delta$. 

Given an $X_r$, for any $\pi \in S_n$, we consider two cases:

\begin{enumerate}
    \item If $\pi \in N_0$, then $d(x_i,x_j) = d(x_{\pi(i)},x_{\pi(j)})$ $\forall$ $(i,j)$. Note that the identity permutation belongs to $N_0$ as:
    \begin{align}
    &D_\pi(X,X_r) 
    =  \frac{1}{2} \max_{i,j}{|d(x_i,x_j)- d(\tilde{x}_{\pi(i)},\tilde{x}_{\pi(j)})|}  \nonumber \\
    =&  \frac{1}{2} \max_{i,j}{|d(x_{\pi(i)},x_{\pi(j)})- d(\tilde{x}_{\pi(i)},\tilde{x}_{\pi(j)})|}  = \frac{1}{2} \max_{i,j}{|d(x_i,x_j)- d(\tilde{x}_i,\tilde{x}_j)|} \label{eq:id}
    \end{align}
    Since $d(x_i,\tilde{x}_i) = d(x_j,\tilde{x}_j) = r$, from Triangle inequality, we have:
    \begin{align*}
        d(\tilde{x}_i,\tilde{x}_j) &\leq d(x_i,x_j) + d(x_i,\tilde{x}_i) +  d(x_j,\tilde{x}_j) = d(x_i,x_j) + 2r \\
        d(\tilde{x}_i,\tilde{x}_j) &\geq d(x_i,x_j) - d(x_i,\tilde{x}_i) -  d(x_j,\tilde{x}_j) = d(x_i,x_j) - 2r
    \end{align*}
    Therefore, for all $i,j$, we have:
    \begin{align}
         |d(\tilde{x}_i,\tilde{x}_j) -  d(x_i,x_j)| \leq 2r \label{eq:xypc}
    \end{align}
    This implies  $D_\pi(X,X_r) \leq r $ for $\pi \in N_0$.
    \item If $\pi \in N_1$, without loss of generality, we assume the pair $(x_1,x_2)$ maximizes $|d(x_i,x_j) - d(x_{\pi(i)},x_{\pi(j)})|$. For convenience, we denote $\pi(1) = 3$ and $\pi(2) = 4$. From the fact that $\pi \in N_1$, we have $| d(x_1,x_2) - d(x_3,x_4)| \geq \delta$. On the other hand, from (\ref{eq:xypc}), $| d(\tilde{x}_3,\tilde{x}_4) - d(x_3,x_4)| \leq 2r$.
    Thus, from Triangle inequality, we obtain:
    \begin{align*}
     | d(\tilde{x}_3,\tilde{x}_4) - d(x_1,x_2)| 
    \geq &  | d(x_1,x_2) - d(x_3,x_4)|  - |d(\tilde{x}_3,\tilde{x}_4) - d(x_3,x_4)|\\
    \geq & \delta - 2r \geq 2r
    \end{align*}
    Since $D_\pi(X,X_r) \geq \frac{1}{2} |d(x_1,x_2) - d(\tilde{x}_3,\tilde{x}_4)|$, we establish
    $ D_\pi(X,X_r) \geq r$ for $\pi \in N_1$.
    \end{enumerate}
From the above analysis, we can conclude $D_\pi(X,X_r) \leq D_\tau(X,X_r)$ for all $\pi \in N_0$ and $\tau \in N_1$. Combining this with (\ref{eq:id}), we have that the identity permutation is the solution of (\ref{eq:dJ}), which proves the Lemma.
\end{proof}

With the Lemma, Theorem \ref{theorem:proj_vs_orth} can be proved by choosing a specific projection perturbation such that its discrete Gromov-Hausdorff distance is always bigger than the upper bound for such distance of any orthogonal perturbations. The proof of the Theorem is shown below:

\begin{proof} \textbf{of Theorem \ref{theorem:proj_vs_orth}.}
Applying Lemma~\ref{lemma:first} to the orthogonal perturbation $Y = X^\perp_r$ and projection perturbation $Z =X^\text{Proj}_r$, and for any $r$ less than or equal to the minimum of the $\epsilon$ corresponding to each perturbation specified in Lemma~\ref{lemma:first}, we obtain:
\begin{align*}
    d_J (X, Y) = \frac{1}{2} \max_{i,j}{|d(x_i,x_j)- d(y_i,y_j)|} \quad \textup{and} \quad
    d_J (X, Z) = \frac{1}{2} \max_{i,j}{|d(x_i,x_j)- d(z_i,z_j)|} 
\end{align*}

From the triangle inequality (similar to how we show Eq. (\ref{eq:xypc}) in the proof of Lemma~\ref{lemma:first}), we have:
\begin{align*}
    \frac{1}{2} \max_{i,j}{|d(x_i,x_j)- d(y_i,y_j)|} < r \quad \textup{and} \quad
    \frac{1}{2} \max_{i,j}{|d(x_i,x_j)- d(z_i,z_j)|} \leq r
\end{align*}
where the first inequality is strict due to the orthogonality of the perturbation.

Given such $r$, consider the following perturbation $Z = X^\text{Proj}_r$ where $x_1, x_2, z_1, z_2$ are collinear and $|d(x_1,x_2) - d(z_1,z_2)| = 2r$, and $z_i = x_i$ for $i = 3, ..., n$. The $d_J$ distance between such $Z$ and $X$ is greater than or equal to $r$, which proves our claim. 
\end{proof}

This theoretical result suggests the following perturbation scheme: Given a manifold embedded in an affine subspace of $\R^N$ and a fixed amplitude $r$ of perturbation, perturbing the manifold in the orthogonal directions with respect to the affine subspace is preferable to random perturbation, as it minimizes the topological difference between the perturbed manifold and the original.

\section{Persistent homology analysis with the Bottleneck distance} \label{sect:bd}

The results from the previous section show that on the worst-case basis, the orthogonal perturbation is preferable to the projection perturbation. However, 
when we apply them to actual datasets, how do they compare on average? Since the discrete Gromov-Hausdorff distance is still 
computationally intractable for a Monte-Carlo analysis, we choose a different approach: \textbf{persistent homology}.

For the last 30 years, there have been new developments in the field of algebraic topology, which was classically very abstract and theoretical, toward real-world applications. The newly discovered field, commonly referred to as \textit{applied topology} or \textit{topological data analysis}, is centered around a concept called \textit{persistent homology}. Interested readers can refer to \citep{ghrist2014elementary, edelsbrunner2010computational, ghrist2008barcode} for an overview of the subject. 

For any topological space, homology is a topological invariant that counts the number of \textit{holes} or \textit{voids} in the space. Intuitively, given any space, the $0^{\text{th}}$-homology counts the number of connected components, the $1^{\text{st}}$-homology counts the number of loops, the $2^{\text{nd}}$-homology counts the number of 2-dimensional voids, and so on. The homology groups of dimension $i$ are denoted by $H_i$.

Given a point cloud sampled from a manifold, we want to recapture the homological features of the original manifold from these discrete points. The idea is to construct a sequence of topological spaces along some \textit{timeline} and track the evolution of the topological features across time. The longer the features \textit{persist} (and hence the name \textit{persistent homology}), the more likely they are the actual features of the original manifold. Given a point cloud $X$ and a dimension $i$, the persistence diagram $D_i(X)$ is the set of points $(b,d) \in \R^2$ corresponding to the birth and death time of these features in the aforementioned timeline.

For two point clouds, and in particular for their two persistence diagrams, there are several notions of distances between them, representing how \textit{similar} they are as topological spaces. The most commonly used distance in practice is the \textbf{Bottleneck distance}. 
\begin{definition}
[Bottleneck distance]
Let $X$ and $Y$ be two persistence diagrams. The Bottleneck distance $W_\infty (X,Y)$ is given by
\begin{align*}
    W_\infty (X,Y) = \inf_{\varphi: X \to Y} \sup_{x \in X} \| x-\varphi(x)\|_\infty
\end{align*}
where the infimum is taken over all matchings $\varphi:X \to Y$  (which allows matchings to points with equal birth and death time).
\end{definition}

For simplicity, we shorthand the Bottleneck distance between persistence diagrams of $X$ and $Y$ in dimension $i$ to $H_i(X,Y)$ instead of $W_\infty(D_i(X), D_i(Y))$. As the notation takes in 2 parameters in $X$ and $Y$, this is not to be confused with the homology group of the specified spaces. Note that two point clouds with small Bottleneck distance can be considered topologically similar. 

The bottleneck distance is highly correlated to the Gromov-Hausdorff distance, as the bottleneck distance of two persistence diagrams of the same dimension is bounded above by their Gromov-Hausdorff distance \citep{chazal2009signatures}:
\[
W_\infty(D_i(X), D_i(Y)) \leq d_GH(X, Y)
\]
for every dimension $i$.

\begin{table}[ht]
\caption{The parameters of the synthetic datasets. The perturbation column shows the average radius of the perturbation applied on each data point. 
}\label{table:syn_params}
\resizebox{0.99\linewidth}{!}{
\begin{tabular}{@{}ccccccc@{}}
\toprule
\textbf{Dataset}  & \textbf{Parameter}    & \textbf{No. points} & \textbf{Data's noise} & \textbf{Perturbation} & \textbf{No. runs} & \textbf{EMaP's dim} \\ \midrule
Line       & Length: 10              & 100                 & 0.1                   & $\approx 0.15$                          & 100               & 1                   \\
Circle     & Radius: 1              & 400                 & 0.1                   & $\approx 0.1$                            & 100               & 2                   \\
2 intersecting circles  & Radius: 1 & 400                 & 0.01                  & $\approx 0.1$                              & 100               & 2                   \\
2 concentric circles & Radius: 1   & 400                 & 0.01                  & $\approx 0.1$                              & 100               & 2                   \\
Spiral          & Radius: [0,2]         & 1000                & 0.02                  & $\approx 0.05$                              & 100               & 2                   \\ \bottomrule
\end{tabular}
}
\end{table}

\begin{table}[ht]
\caption{The parameters of the real-world datasets.}\label{table:real_params}
\centering
\resizebox{0.9\linewidth}{!}{
\centering
\begin{tabular}{@{}cccccc@{}}
\toprule
\textbf{Experiment} & \textbf{No. data points} & \textbf{No. feats} & \textbf{Feature's values} & \textbf{No. runs} & \textbf{EMaP's dim} \\ \midrule
COMPAS              & 7214                     & 100                & \{0,1\}                      & 100               & 2                   \\
German Credit       & 1000                     & 28                 & \{0,1\}                      & 100               & 2                   \\
CC                  & 2215                     & 100                & \{0,1\}                      & 100               & 2                   \\  \midrule
MNIST               & 60000                    & $28\times28$       & {[}0,1{]}                                                                                                   & 100               & 2 and 3                   \\ 
Fashion-MNIST               & 60000                    & $28\times28$       & {[}0,1{]}                                                                                                   & 100               & 2 and 3  \\ \bottomrule
\end{tabular}
}
\end{table}

\textbf{Monte-Carlo simulations.} The Bottleneck distance is much more calculable than the Gromov-Hausdorff distance, and there are available software packages depending on the use cases. As such, we run Monte-Carlo simulations to compute the Bottleneck distances on 5 synthetic datasets, 3 real-world tabular datasets and 2 real-world image datasets to confirm our hypothesis that the orthogonal perturbation preserves the topology better than the projection perturbation \textit{on average}. The synthetic datasets are some noisy point clouds of certain 2-dimensional shapes in 3-dimensional space. The tabular datasets are the COMPAS~\citep{compassdataset}, German Credit~\citep{germancredit},  and Communities and Crime~\citep{communitycrime}. The image datasets are MNIST~\citep{lecun2010} and Fashion-MNIST~\citep{Xiao2017FashionMNISTAN}. Table~\ref{table:syn_params} and \ref{table:real_params} provide more details about those datasets. We use the Ripser Python library~\citep{ctralie2018ripser} to compute the Bottleneck distances in our experiments. All reported Bottleneck distances are normalize with the noise added to the point clouds for more intuitive visualization.


\begin{table}[ht]
\centering
\caption{The normalized $H_0$ and $H_1$ Bottleneck distances for Gaussian (G), projection (P), and orthogonal (O) perturbations on synthetic datasets. Visualizations of the actual perturbations are provided in Appendix~\ref{appendix:viz}.}
\label{table:h0h1syn}
\resizebox{1.0\linewidth}{!}{
\begin{tabular}{ccccccccccc} 
\toprule
\multirowthead{-4}{\textbf{Data}}  & \multicolumn{2}{c}{\imgline} & \multicolumn{2}{c}{\imgcircle} & \multicolumn{2}{c}{\imgcircletwo} & \multicolumn{2}{c}{\imgcirclethree}                  & \multicolumn{2}{c}{\imgspiral}                   \\ 
\hline
\multirowthead{-6}{\textbf{$H_0$}} & \multicolumn{2}{c}{\imgHzline}   & \multicolumn{2}{c}{\imgHzCircle}   & \multicolumn{2}{c}{\imgHzCircleii}   & \multicolumn{2}{c}{\imgHzCircleiii}  & \multicolumn{2}{c}{\imgHzSpiral}  \\ 
\hline
\multirowthead{-6}{\textbf{$H_1$}} &   &  & \multicolumn{2}{c}{\imgHOneCircle}   & \multicolumn{2}{c}{\imgHOneCircleii}   & \multicolumn{2}{c}{\imgHOneCircleiii}  & \multicolumn{2}{c}{\imgHOneSpiral}  \\ 
\bottomrule
\end{tabular}
}
\end{table}

Table~\ref{table:h0h1syn} reports the means $H_0$ and $H_1$ Bottleneck distances of the perturbations on the synthetic datasets. The number of data points and the noise's level are chosen mainly for nice visualizations (shown in Appendix~\ref{appendix:viz}). The results show that orthogonal perturbation consistently results in lower $H_0$ distances for line-shaped dataset and lower $H_1$ distances for cycle-shaped datasets. Note that in general, $H_1$ is the better topological indicator for cycle-shaped datasets compared to $H_0$, since cycles or holes (detected by $H_1$) are harder to replicate than connected components (detected by $H_0$). 

\begin{figure}[ht]
     \centering
         \includegraphics[width=0.85\textwidth]{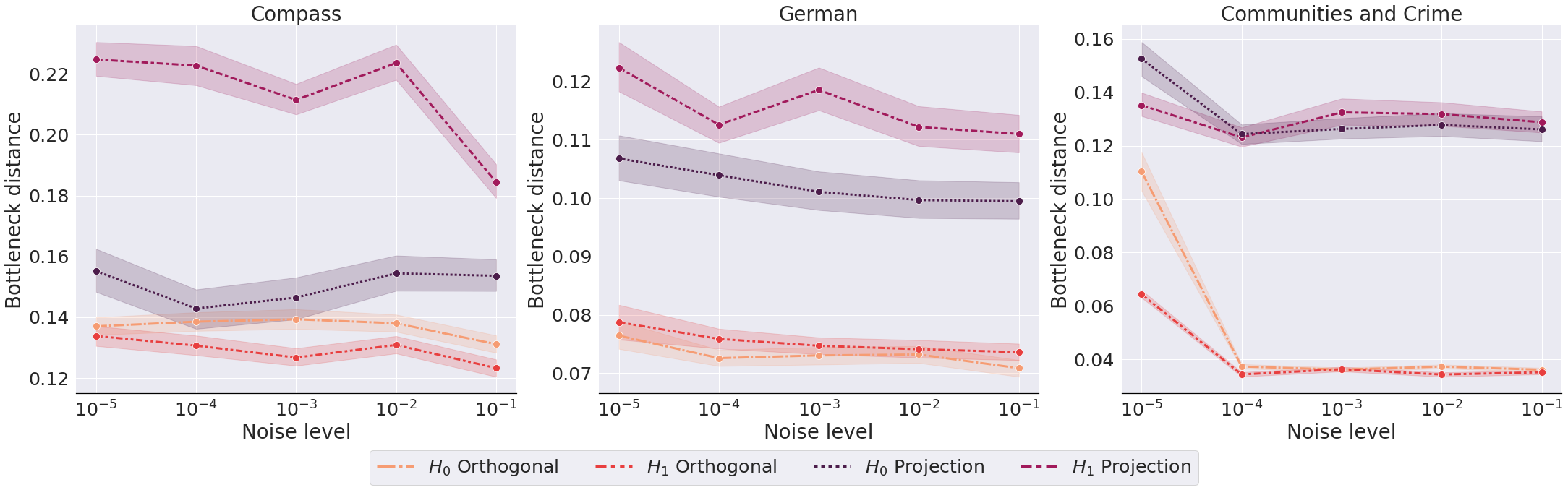}
     \caption{The normalized $H_0$ and $H_1$ Bottleneck distances for orthogonal and projection perturbations on 3 real-world datasets at different noise levels. The x-axis shows the average perturbation's radius applied on each data point (log-scale).}\label{fig:h0h1}

\vspace{4mm}
 
     \centering
         \includegraphics[width=\textwidth]{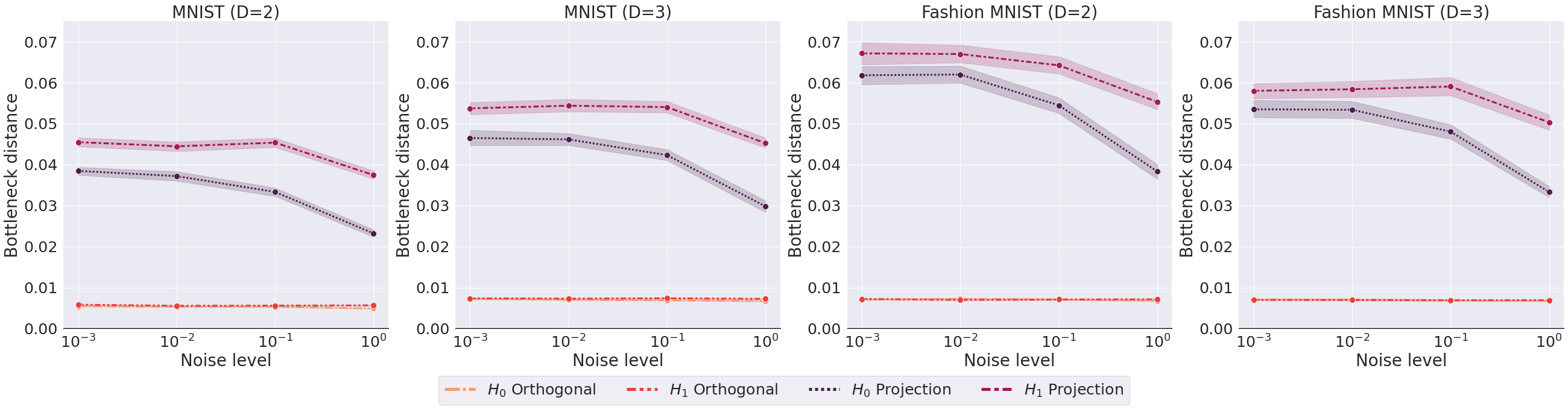}
     \caption{The normalized $H_0$ and $H_1$ Bottleneck distances for orthogonal and projection perturbations on 2 image datasets at different noise levels. The x-axis shows the average perturbation's radius applied on each data point (log-scale).}\label{fig:h0h1_mnist}
 \end{figure}

 For the real-world dataset, we conduct the experiments with perturbations of different noise levels and report results in Fig.~\ref{fig:h0h1} and \ref{fig:h0h1_mnist}. It can be observed that both $H_0$ and $H_1$ Bottleneck distances of the persistence diagrams of the orthogonal perturbation are significantly smaller than those of the projection perturbation on all experiments.

\section{EMaP Algorithm} \label{sect:Emap}

The ideal perturbations for perturbation-based explainers are those drawn from the data distribution since models are trained to operate on that distribution~\citep{Marco2016,Ribeiro2018AnchorsHM}. However, most perturbation schemes ignore the data distribution in the process of generating the perturbations 
(see Sect.~\ref{sect:related}). Furthermore, predictions on the perturbations do not necessarily hold local information about the explained inputs, which is one main reason for the usage of some distance or kernel functions measuring how similar the perturbations are to the explained input. Note that those distances and kernels are normally functions of other distances such as $L_1$ and $L_2$ in the input space $\R^N$, which might not correctly capture the notion of similarity. 

\begin{figure}[ht]
     \centering
         \includegraphics[width=0.56\linewidth]{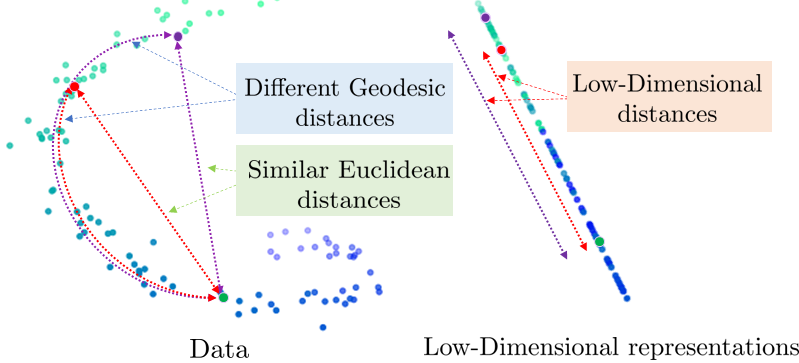}
     \caption{Euclidean distances computing in the input space might not capture the actual distances between the data points (left). Distances in low-dimensional space can help with the issue (right).}
     \label{fig:geo_dist}
\end{figure}

By operating in the low-dimensional manifold, EMaP  can overcome those issues. First, if topological similarity implies similarity in the model's predictions, maintaining the topological structure of the original data should improve the relevance of the model's predictions on the perturbations. Therefore, explaining the model with orthogonal perturbations, which helps preserve the topology better, should be more beneficial. Furthermore, the manifold provides a natural way to improve the similarity measurement among data points. Fig.~\ref{fig:geo_dist} shows the issue of similarity measurement based on Euclidean distance in the input space $\R^N$. As the distance ignores the layout of the data, further points on the manifold might result in the same similarity measure. On the other hand, the low-dimensional distances computed on the manifold take into account the layout and can overcome that issue.

\textbf{Algorithm overview.}
Given an input to be explained, the output of EMaP are the perturbations along the manifold's orthogonal directions $X_r$ and  their low-dimensional distances  $D_r$ to that input. 
The pseudo-code of EMaP is shown in Alg.~\ref{alg:init}.  The first step is to learn an embedding function, i.e. a \textit{mapper}, transforming the data to the low dimension (line 2). Then, $p$ samples from each label are selected and combined with the explained input $x_0$ into a set, called the \textit{pivots} (line 3 to 7). After that, EMaP generates perturbations along the orthogonal directions of the manifold from each pivot (line 10). The usage of pivots is to provide the explainer a wider range of perturbations for better performance. In the next paragraphs, we will describe those key steps in more details. 

\begin{algorithm}[ht]
\caption{EMaP}
\label{alg:init}
\textbf{Input}: Data to explain $x_0$, a subset of training data $(X,y)$, number of pivots per labels $p$, number of perturbations per pivot $k$, lower dimension $V$ and noise level $r$.\\
\textbf{Output}: $X_r$ and $D_r$. $X_r$ contains $k(pl + 1)$ orthogonal perturbations locally around $x_0$ and points in $X$. $D_r$ contains the low-dimensional distances of points in $X_r$ to $x_0$ ($l$ is the number of unique labels in $y$). 

\begin{algorithmic}[1] 
\STATE Initialized an EMaP sampler object $\mathcal{M}$.
\STATE $\mathcal{M}$.mapper $\leftarrow$ Mapper to the manifold of dimension $V$ of $X$
\STATE $\mathcal{M}$.pivots $\leftarrow \emptyset$
\STATE Include $x_0$ to $\mathcal{M}$.pivots
\FOR {each class $l$ in $y$}
    \STATE Include $p$ samples of class $l$ to $\mathcal{M}$.pivots
\ENDFOR

\STATE $X_r\leftarrow \emptyset$, $D_r\leftarrow \emptyset$
\FOR {each data point $x$ in $\mathcal{M}$.pivots }
    \STATE $\Tilde{X},\Tilde{D} \leftarrow \mathcal{M}.\textup{gen\_perturbation}(x,k,r)$
    \STATE Include $\Tilde{X}$ to $X_r$ and include $\Tilde{D}$ to $D_r$
\ENDFOR

\STATE \textbf{return} $X_r$ and $D_r$.
\end{algorithmic}
\end{algorithm}

\textbf{The mapper.} In EMaP, the mapper is learnt from a manifold approximated by UMAP~\citep{mcinnes2018umap-software}. 
Since the manifold learnt by UMAP is optimized for global information, the orthogonal directions computed on top of that manifold at local data points are prone to high error. This can degrade the correctness of orthogonal perturbations significantly. To overcome this issue, EMaP learns a local-subspace for each pivot and generates the orthogonal perturbations on top of that subspace. Intuitively, the local-subspace is a local affine approximation of the manifold. Therefore, the resulted orthogonal directions are more finely tuned for the local points. We denote $G_x$ as the $N \times V$ matrix characterizing the local-subspace at $x$. Since $G_x$ is a linear approximation of data points near $x$, by denoting  $\omega: \mathbb{R}^N \rightarrow \mathbb{R}^V$ as the function embedding input data to the manifold. By denoting $ x^{\textup{low}} =\omega(x)$, we have $z^{\textup{low}} \approx G_x^\top z$ and $  z \approx G_x \ z^{\textup{low}}$,
where $z \in \R^N$ are points near $x$ and $z^{\textup{low}}$ are their embedding in $\R^V$. In our current implementations, the set of pivots contains the explained data point and $p$ data points sampled from each class label $l$ (see Algorithm~\ref{alg:init}).

\textbf{EMaP orthogonal perturbations.} The key step of EMaP is the generation of orthogonal perturbations $\Tilde{x}$ from $x$ (line 10, Alg.~\ref{alg:init}), which can be described by the  following equation:
\begin{align}
    & \Tilde{x} = x + \textup{noise} - \textup{Proj}_{G_x}(\textup{noise}) \label{eq:gen_perturb}
\end{align}
where the noise is sampled from a multivariate normal distribution and $ \textup{Proj}_{G_x}(\textup{noise})$ is the projection of the noise  on the local-subspace characterized by $G_{x}$. Upon obtaining the orthogonal perturbations, we can compute their low-dimensional embedding using the mapper transform function $\omega$. The pseudocode for this computation is in Algorithm~\ref{alg:gen_perturb}.

\begin{algorithm}[ht]
\caption{gen\_perturbation}
\label{alg:gen_perturb}
\textbf{Input}: Input $x$, number of perturbation $k$ and noise level $r$. \\
\textbf{Output}: $k$ orthogonal perturbations of $x$ and their low-dimension distances to $x$.

\begin{algorithmic}[1] 
\STATE $G_x \leftarrow$ self.get\_local\_subspace($x$) 
\STATE $\Tilde{X} \leftarrow \emptyset$
\FOR{$1\leq i \leq k$:}
    \STATE $\textup{noise} \leftarrow \mathcal{N}(0,\Sigma_r)$
    \STATE $\Tilde{x} \leftarrow x + \textup{noise} - \textup{Proj}_{G_x}(\textup{noise})$
    \STATE Include $\Tilde{x}$ into $\Tilde{X}$
\ENDFOR
\STATE $\Tilde{X}^{\textup{low}} \leftarrow \textup{self.mapper.transform}(\Tilde{X})$
\STATE $x^{\textup{low}} \leftarrow \textup{self.mapper.transform}(x)$
\STATE 
$\Tilde{D} \leftarrow$ distances of each member in $\Tilde{X}^{\textup{low}}$ to $x^{\textup{low}}$
\STATE \textbf{return} $\Tilde{X}$ and $\Tilde{D}$
\end{algorithmic}

\end{algorithm}

\textbf{Local-subspace approximation.} The correctness of the orthogonal perturbations, i.e. whether the perturbations are actually lying in the orthogonal subspace, is heavily dependent on the correctness of the computation of $G_{x}$. We now discuss how EMaP learns the local-subspace matrix $G_x$. 

Ideally, given a set of data $Z$ near $x$ in the manifold, we can solve the following optimization for $G_x$ :
\begin{align}
    G_x &= \arg \min_G  \sum_{z\in Z}\|  G \omega(z) - z\|, \label{eq:idealGx}
\end{align}
where $\omega$ is the transform function embedding input data to the manifold learnt in the previous step.
An intuition is, for all $z$ in the manifold and near $x$, we are searching for a matrix $G$ such that the inverse mapping $G \omega(z) \approx  G z^{\textup{low}}$ approximately equals its original value $z$. Note that if the embedding $\omega$ is exactly on $Z$ and the manifold is affine, the optimization (\ref{eq:idealGx}) can achieve its optimal value $0$ for some $G$. Since it is not trivial to obtain the set $Z$ belonging to the manifold, EMaP perturbs around $x$ with some random noise, and solves the following the optimization instead:
\begin{align}
    \hat{G}_x &= \arg \min_G \sum_{r \in B} \|  G \omega(x + r) - (x + r)\|, \label{eq:emap_local_hp}
\end{align}
where $B$ is a ball centering at $0$ with noise radius $r$. EMaP solves (\ref{eq:emap_local_hp}) for an approximation of the local-subspace. Further details of this step are provided in Algorithm~\ref{alg:get_local}.

\begin{algorithm}[ht]
\caption{get\_local\_subspace}
\label{alg:get_local}
\textbf{Input}: Data point $z$.\\
\textbf{Hyper-parameters}: Number of training samples $k_T$ and noise level for training $r_T$.\\
\textbf{Output}: Matrix $G$ characterize the local-subspace at $z$

\begin{algorithmic}[1] 
\STATE $\Tilde{Z} \leftarrow \emptyset$
\FOR{$1\leq i \leq k_T$:}
    \STATE $\textup{noise} \leftarrow \mathcal{N}(0,\Sigma_{r_T})$
    \STATE $\Tilde{z} \leftarrow z + \textup{noise}$
    \STATE Include $\Tilde{z}$ into $\Tilde{Z}$
    \ENDFOR
    \STATE $\Tilde{Z}^{\textup{low}} \leftarrow \textup{self.mapper.transform}(\Tilde{Z})$
    \STATE $G \leftarrow \arg\min_W \| \Tilde{Z} - W\Tilde{Z}^{\textup{low} } \|_2$
\STATE \textbf{return} $G$
\end{algorithmic}
\end{algorithm}

We now discuss the gap 
between the ideal solution $G_x$ and the approximation $\hat{G}_x$ used by EMaP, i.e. the solution of (\ref{eq:idealGx}) and (\ref{eq:emap_local_hp}).
This can be characterized by bounding the error between $\{G_x \omega(z)\}$ and $\{\hat{G}_x \omega(z)\}$, i.e. the reconstructed signals in $\R^N$ by using ${G}_x$ and $\hat{G}_x$, respectively. Lemma~\ref{lemma:computationalbound} provides a bound on that reconstruction error where the set $Z$ in (\ref{eq:idealGx}) is the projection of a ball $B$ on the data manifold. The bound holds under a mild assumption that the optimal objective of (\ref{eq:idealGx}) is not larger than that of (\ref{eq:emap_local_hp}). We find this assumption reasonable and intuitive: as the set $Z$ is in a subspace of dimension $V$ and the set of $x+r$ is a ball in $\R^N$, finding a subspace of dimension $V$ approximating the subspace containing $Z$ should give a much lower error.

\begin{lemma} \label{lemma:computationalbound}
Assume that all data points $x$ belong to the same affine space $\R^V$. Let $\text{Proj}$ be the projection onto $\R^V$, then under the above assumption on the optimization (\ref{eq:idealGx}) and (\ref{eq:emap_local_hp}), the reconstruction error on perturbed data points is upper bounded by:
\begin{align*}
    & \| \hat{G}_x \omega(x+r) -  G_x \omega (\textup{Proj}(x+r)) \| 
    \leq F_B(\omega) + \| r^{\perp}\|,
\end{align*}
where $r^{\perp}$ is the orthogonal components of $r$ and $F_B(\omega) := \min_G  \sum_{r\in B}\|  G \omega(x+r) - (x+r)\|$. 
\end{lemma}

\begin{proof}
For simplicity, we rewrite:
\begin{align*}
    G_1 &= \arg \min_G  \sum_{r\in B}\|  G \omega(x+r) - (x+r)\|, \\
    G_2 &= \arg \min_G  \sum_{r\in B}\| G \omega(\textup{Proj}(x + r)) - \textup{Proj}(x + r)\|
\end{align*}

The assumption regarding the objectives (\ref{eq:idealGx}) and (\ref{eq:emap_local_hp}) mentioned in the Lemma can be rewritten as:
\begin{align*}
    \sum_{r\in B}\| G_2 \omega(\textup{Proj}(x + r)) - \textup{Proj}(x + r)\| \leq \sum_{r\in B}\|  G_1 \omega(x+r) - (x+r)\|
\end{align*}
We find this assumption reasonable since its left hand side is equal to $0$ in the ideal scenario, i.e. $x \in \R^V$ for all $x$ in our dataset. With that, we have:
\begin{align*}
    &2 \| G_1 \omega(x+r) - (x + r) \| \\
    \geq& \| G_1 \omega(x+r) - (x + r) \| +
    \| G_2 \omega(\textup{Proj}(x + r)) - \textup{Proj}(x + r) \| \\
    \geq& \| G_1 \omega(x+r) - G_2 \omega(\textup{Proj}(x + r))  -
    (x + r - \textup{Proj}(x + r) ) \| 
    \\
    \geq& \| G_1 \omega(x+r) - G_2 \omega(\textup{Proj}(x + r)) || -
     || (x + r) - \textup{Proj}(x + r) ) \| \\
     = & \| G_1 \omega(x+r) - G_2 \omega(\textup{Proj}(x + r)) || - \| r - \textup{Proj}(r) \|,
\end{align*}
where the last two inequalities are from the Triangle Inequality. The last equality is due to the fact that $(x + r) - \textup{Proj}(x + r) $ is the orthogonal components of $x + r$ and that $x$ has no orthogonal components.
\end{proof}

Note that $F_B(\omega)$ is small if the manifold is affine in the neighborhood $B$ of $x$ and $\omega$ is good, i.e. $\hat{G}_x$ is a good estimator for $G_x$ under the above assumption. 



\section{Experiments} \label{sect:exp}

We evaluate EMaP on two main objectives: explainer's performance and perturbations' robustness. Our experiments are conducted on 3 tabular datasets, 2 images datasets, and 4 text datasets of reviews in multiple domains. They are COMPAS~\citep{compassdataset},
Communities and Crime~\citep{communitycrime}, German Credit~\citep{germancredit}, MNIST~\citep{lecun2010}, Fashion-MNIST~\citep{Xiao2017FashionMNISTAN}, and 4 reviews datasets in the Multi-Domain Sentiment~\citep{multibook}. 

\textbf{Dataset, models and explainer's hyper-parameters.} All reported results for real-world datasets include at least 2000 data points and 100 runs, except for those of German Credit where the data only consists of 1000 samples. 

The experimental models for the text dataset is the $L1$ logistic regression implemented by the LIME paper~\citep{Marco2016} with the ground-truth sets of explanatory features. The model of testing for the two image datasets are 2-layer convolutional networks implemented in Pytorch~\citep{pytorch} with test set accuracy of 98\% for MNIST and 93\% for Fashion MNIST.

The model's inputs of all experiments are normalized between 0 and 1. The noise vector used to generate perturbations has radius $10^{-3}$ for text data and $10^{-4}$ for image data, which are in the range shown in the previous experiments in Fig.~\ref{fig:h0h1} and \ref{fig:h0h1_mnist}. The noise radius $r_T$ used to approximate the local-subspaces (Algorithm~\ref{alg:get_local}) is chosen equal to the noise radius $r$ for perturbation. The selection of these radii and the low-dimensional value $V$ of the manifold depends on the dataset. For UMAP's hyper-parameters, we use their default settings with n\_components $=2$ and min\_dist $=0.1$. Our source code is attached in the supplementary material of this submission. Finally, for fair comparison, the number of perturbations used to generate the explanation of any reported methods is 1000.

\textbf{Technical implementation of EMaP and baselines.} 
Generally, EMaP perturbations can be used to leverage any model-agnostic perturbation-based methods; however, the modification may require significant changes on the existing implementations of the methods. In our experiments, we use EMaP to leverage LIME~\citep{Marco2016} as a proof of work, i.e. we show that EMaP can improve LIME in term of performance. We use notation EMaP to indicates \textit{LIME with EMaP's perturbations} in our following experimental results. We choose LIME to demonstrate the advantages of EMaP since it requires few changes to integrate EMaP's perturbations. This helps demonstrate fairly the gain of applying EMaP on explanation methods. 

We now explain in more details how we leverage LIME with EMaP. As described in the original paper, LIME's explanation is the solution of the following optimization:
\begin{align*}
    \argmin_{g \in \mathcal{G}} \mathcal{L} (f,g, \pi_x) + \Omega(g)
\end{align*}
where $\mathcal{L}$ is a loss between the explained function $f$ and the explanation function $g$, $\mathcal{G}$ is the class of linear models and $\pi_x$ is an exponential kernel defined on some distance function (see~\citep{Marco2016} for more details). To use EMaP, we set the loss function as:
\begin{align*}
    \mathcal{L} (f,g, \pi_x) = \sum_{\tilde{x} \in X_r} \pi_x(\tilde{x}) (f(\tilde{x}) - g(x - \tilde{x}))
\end{align*}
where $\pi_x(\tilde{x}) = \exp(-\tilde{D}(x,\tilde{x})^2/\sigma^2)$ with the distance $\tilde{D}$ computed as in Algorithm~\ref{alg:gen_perturb} and $g$ is the linear function of the changes in each input's features. 

Throughout our experiments, we compare EMaP-LIME (or EMaP for short) to LIME with different perturbation schemes. Our goal is to demonstrate the advantage of using EMaP to generate explanation. We also include experimental results of some other black-box and white-box explanation methods for comparison. Specifically, we include results of following methods:
\begin{itemize}
    \item LIME zero: LIME with perturbations whose perturbed features are set to zero. This method is used by LIME in explaining text data.
    \item LIME+: LIME with perturbations whose perturbed features are added with Gaussian noise. Ths method is used by LIME in explaining image data.
    \item LIME*: LIME with perturbed whose perturbed features are multiplied with uniform noise between 0 and 1. This can be considered as a smoother version of LIME zero.
    \item KernelSHAP: a black-box method based on Shapley value~\citep{Scott2017}, whose perturbed features are set to average of some background data.
    \item GradientSHAP: a white-box method based on Shapley value~\citep{Scott2017}, which relies on the gradient of the model.
    \item DeepLIFT: a white-box method based on back-propagating the model~\citep{Avanti2017}.
\end{itemize}

\begin{figure}[ht]
	\centering
	\begin{subfigure}{0.48\linewidth}
     \centering
         \includegraphics[width=0.9\linewidth]{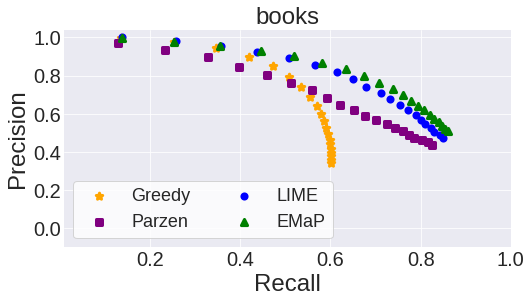}
	\end{subfigure}
	\begin{subfigure}{0.48\linewidth}
     \centering
         \includegraphics[width=0.9\linewidth]{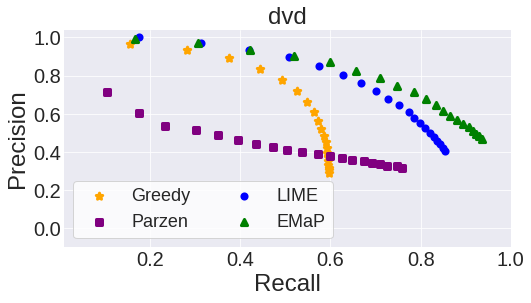}
	\end{subfigure}
	\begin{subfigure}{0.48\linewidth}
     \centering
         \includegraphics[width=0.9\linewidth]{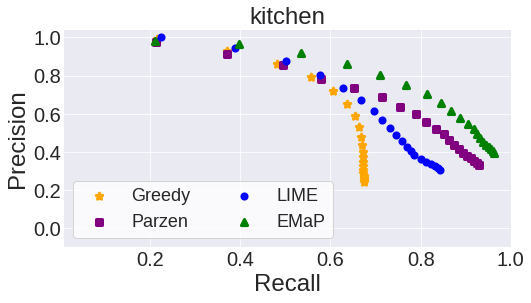}
	\end{subfigure}
	\begin{subfigure}{0.48\linewidth}
     \centering
         \includegraphics[width=0.9\linewidth]{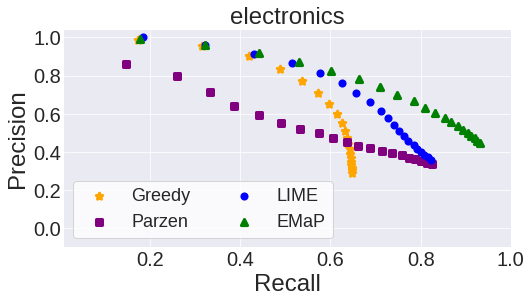}
	\end{subfigure}
	\caption{The precision and recall of explanations returned by Greedy, LIME, Parzen and LIME-EMaP (the higher the better). The dots are in the increasing order of the number of features in explanations (left to right).}
	\label{fig:multi}
\end{figure}

\textbf{Explainer's performance.} 
We first report our experimental result for the sentiment classification task in the Multi-Domain Sentiment dataset~\citep{multibook}. We follow the experimental setup in~\citep{Marco2016}, in which the ground-truth explanatory features of the $L1$ logistic regression model are known. The impact of EMaP's perturbations on the explainer is evaluated by the precision and the recall rate of the features in the explanations. Intuitively, more features in the explanation will increase the recall and decrease the precision. 

Fig.~\ref{fig:multi} shows the scatter plot of precision vs. recall of LIME and LIME with EMaP on 4 review datasets of \textit{books}, \textit{dvds}, \textit{kitchen} and \textit{electronics}. We also provide the result of the Greedy and Parzen explanation methods~\citep{parzen}. In Greedy, the features contributing the most to the predicted class are removed until the prediction changes. On the other hand, Parzen approximates the model globally with Parzen windows and the explanation is the gradient of the prediction. The results clearly show that EMaP consistently improves the faithfulness of LIME.

\begin{figure}[ht]
	\centering
	\begin{subfigure}{0.49\linewidth}
     \centering
         \includegraphics[width=0.96\linewidth]{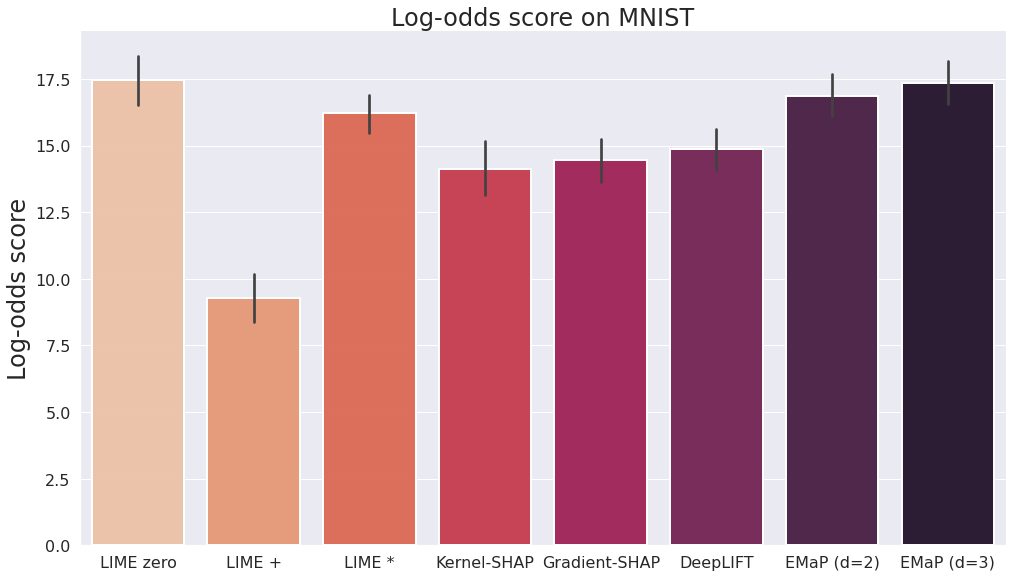}
	\end{subfigure}
	\begin{subfigure}{0.49\linewidth}
     \centering
         \includegraphics[width=0.96\linewidth]{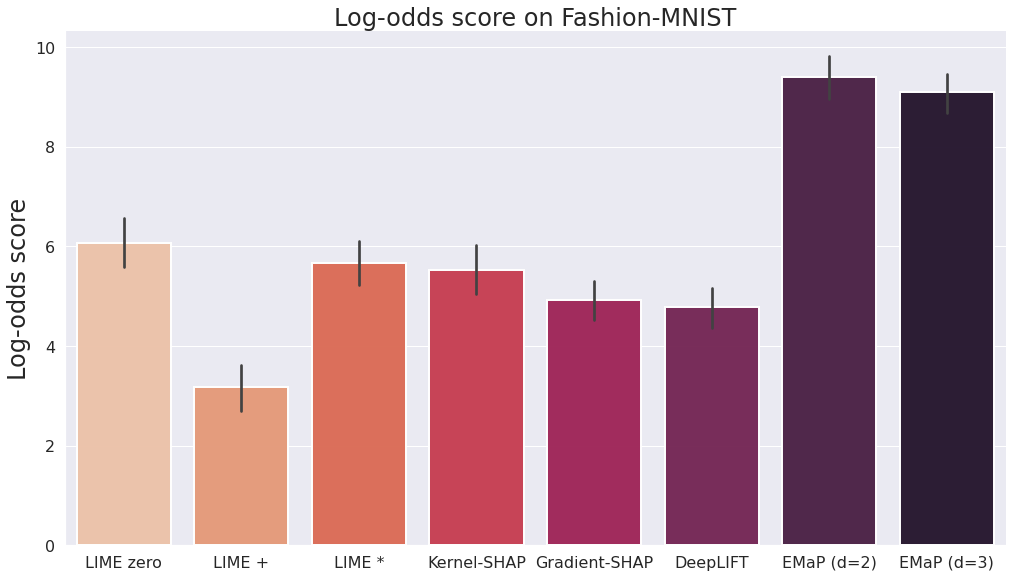}
	\end{subfigure}
	\caption{Log-odds scores of different perturbation-based methods on MNIST and Fashion-MNIST (the higher the better).}
	\label{fig:logodd}
\vspace{4mm}
	\centering
	\begin{subfigure}{0.49\linewidth}
     \centering
         \includegraphics[width=0.96\linewidth]{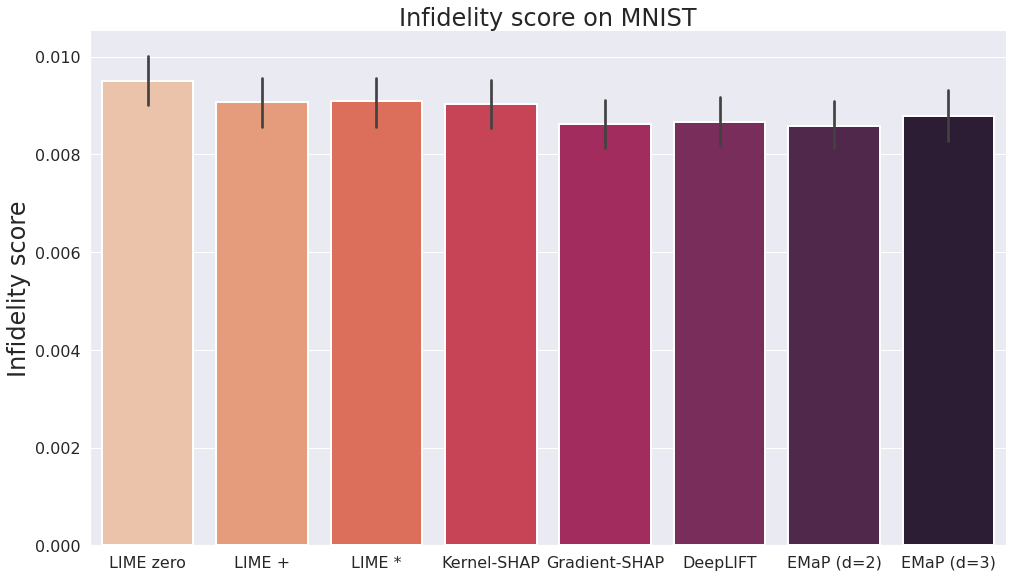}
	\end{subfigure}
	\begin{subfigure}{0.49\linewidth}
     \centering
         \includegraphics[width=0.96\linewidth]{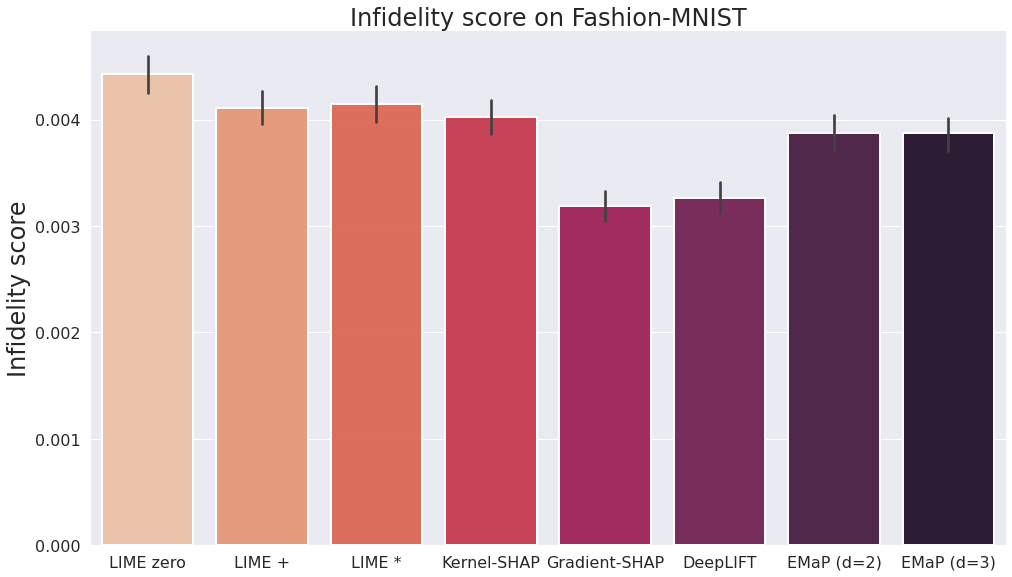}
	\end{subfigure}
	\caption{Infidelity scores of different perturbation-based methods on MNIST and Fashion-MNIST (the lower the better).}
	\label{fig:infidelity}
\end{figure}


Since there is no ground-truth explanations for the MNIST and Fashion MNIST image datasets, we evaluate explanations using the log-odds scores~\citep{Avanti2017} and the infidelity scores~\citep{Yeh2019OnT}. Given an input image and the importance weights of its features, the log-odds score measures the difference between the image and the modified image whose pixels are erased based on their importance weights. In our experiments, the erased pixels are those with the top 20\% weights. Intuitively, the higher the log-odds score, the better the explanation. On the other hand, the infidelity score measures the expected error between the explanation multiplied by a meaningful perturbation and the differences between the predictions at its input and at the perturbation. The metric can be considered as a generalized notion of Sensitivity-$n$~\citep{ancona2018towards}. Intuitively, explanations with lower infidelity are more desirable.

Fig.~\ref{fig:logodd} shows the log-odds scores of EMaP with low-dimension $d=2$ and $d=3$, along with other explanation methods and other perturbation schemes. In MNIST, we can see that EMaP does not degrade the explainer performance compared to LIME in term of log-odds(note that the default setting for LIME in image is LIME+). For Fashion MNIST, EMaP improves the log-odds significantly. Fig.~\ref{fig:infidelity} shows the infidelity scores. It is clear that EMaP has the lowest infidelity score among all black-box methods. Even though the white box methods, KernelSHAP and DeepLIFT, have more information on the explained models than EMaP, they can only outperform EMaP in Fashion-MNIST. Some virtualization of EMaP and other explanation methods are provided in Appendix~\ref{appendix:viz}.

\begin{figure}[ht]
	\centering
	\begin{subfigure}{0.49\linewidth}
     \centering
         \includegraphics[width=0.75\linewidth]{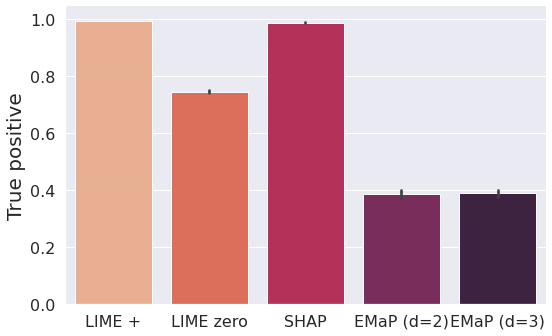}
	\end{subfigure}
	\begin{subfigure}{0.49\linewidth}
     \centering
         \includegraphics[width=0.75\linewidth]{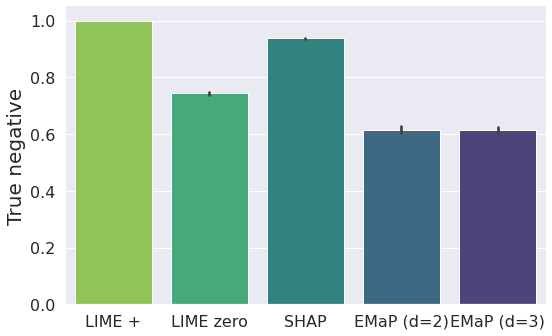}
	\end{subfigure}
	\caption{True-Positive and True-Negative rates of the discriminators on perturbations of different methods on MNIST dataset (the lower the better).}
	\label{fig:disc_mnist}

	\centering
	\begin{subfigure}{0.49\linewidth}
     \centering
         \includegraphics[width=0.75\linewidth]{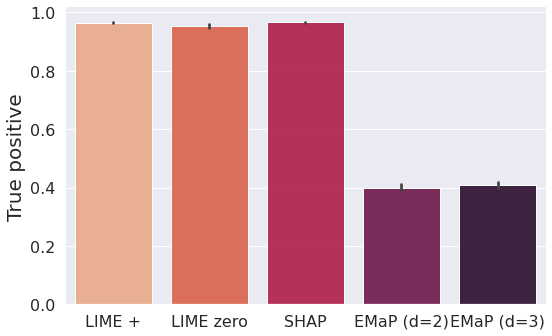}
	\end{subfigure}
	\begin{subfigure}{0.49\linewidth}
     \centering
         \includegraphics[width=0.75\linewidth]{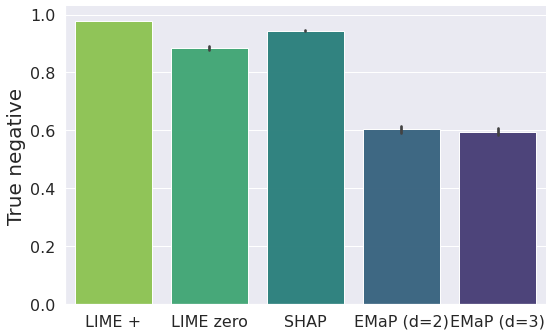}
	\end{subfigure}
	\caption{True-Positive and True-Negative rates of the discriminators on perturbations of different methods on Fashion-MNIST dataset (the lower the better).}
	\label{fig:disc_fashion}

	\centering
	\begin{subfigure}{0.49\linewidth}
     \centering
         \includegraphics[width=0.75\linewidth]{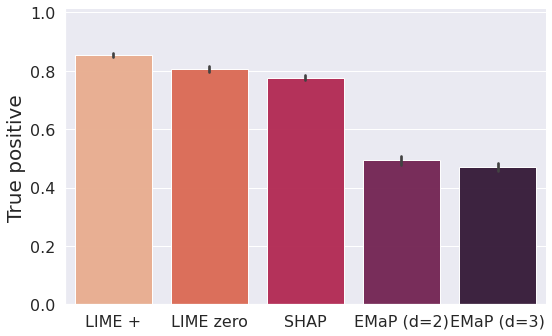}
	\end{subfigure}
	\begin{subfigure}{0.49\linewidth}
     \centering
         \includegraphics[width=0.75\linewidth]{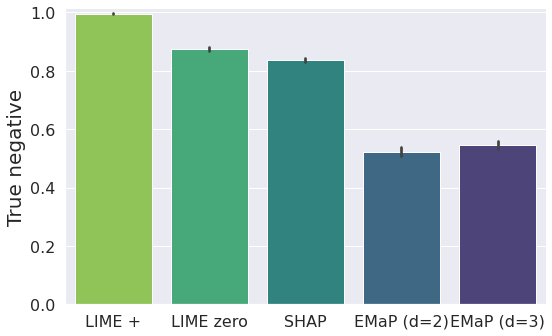}
	\end{subfigure}
	\caption{True-Positive and True-Negative rates of the discriminators on perturbations of different methods on Communities and Crime dataset (the lower the better).}
	\label{fig:disc_cc}

	\centering
	\begin{subfigure}{0.49\linewidth}
     \centering
         \includegraphics[width=0.75\linewidth]{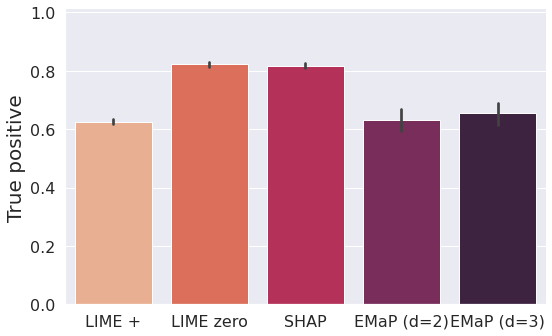}
	\end{subfigure}
	\begin{subfigure}{0.49\linewidth}
     \centering
         \includegraphics[width=0.75\linewidth]{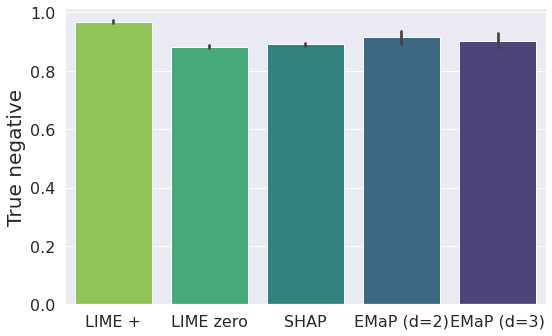}
	\end{subfigure}
	\caption{True-Positive and True-Negative rates of the discriminators on perturbations of different methods on German Credit dataset (the lower the better).}
	\label{fig:disc_german}
\end{figure}



\textbf{Perturbation's robustness.} 
We evaluate the robustness of the perturbation scheme based on the discriminator's performance in differentiating perturbations from the original data. Following the setup in~\citep{FoolingLIMESHAP}, the discriminator is trained with a full-knowledge on explainer's parameters. This discriminator has been shown to be able to recognize perturbations of LIME and SHAP explainers. 

Our experimental results show that EMaP's perturbation is more robust to the discriminator. Specifically, Figs.~\ref{fig:disc_mnist}, \ref{fig:disc_fashion}, \ref{fig:disc_cc} and~\ref{fig:disc_german} show the True-Positive (TP) and True-Negative (TN) rates of discriminators on perturbations of 2 image datasets and 2 tabular datasets. Note that the TP and TN rates are ideally around $50\%$ since they would indicate that the discriminators cannot recognize both the original data and the perturbations. For image datasets, the discriminators can easily recognize perturbations generated by LIME and SHAP. On the other hand, EMaP perturbations significantly lower the success rates of discriminator in recognizing the perturbations. While the explainers' perturbation schemes show to be slightly more robust in the tabular datasets compared to the image datasets, EMaP still improves the  perturbations' robustness remarkably. 

\textbf{Computational resource and run time.} 
Our experiments are conducted on a single GPU-assisted compute node that is installed with a Linux 64-bit operating system. The allocated resources include 32 CPU cores (AMD EPYC 7742 model) with 2 threads per core, and 100GB of RAM. The node is also equipped with 8 GPUs (NVIDIA DGX A100 SuperPod model), with 80GB of memory per GPU. 

The run time of EMaP is mostly dictated by the learning of the embedding function (line 2 of Algorithm~\ref{alg:init}). That initilization step in the tabular dataset for about 2000 data points takes less than  2 minutes. It takes between 240 and 260 seconds for all images of 60000 MNIST/Fashion-MNIST images. 
Given the local-subspaces, the generation of 1000 orthogonal perturbations takes about 0.3 second. Note that the manifold and local-subspaces can be computed before deployment since it does not depends on the explained inputs. For a rough comparison, the processing of perturbations by LIME on 1000 perturbations of both image datasets on the 2-layer network also takes about 0.3 seconds. Thus, the overhead of EMaP at deployment is reasonable. Table~\ref{table:runtime} report the actual run time of EMaP and LIME in the image datasets.

\begin{table}[ht]
\centering
\caption{Run time (in seconds) of EMaP compared to LIME. The reported numbers are \textit{seconds} (for initialization column) and\textit{ seconds per explanation} (for other columns).} \label{table:runtime}
\begin{tabular}{@{}ccccc@{}}
\toprule
              & EMaP Initialization & LIME  & EMaP (d=2) & EMaP (d=3) \\ \midrule
MNIST         & 240-260    & 0.763 & 1.311      & 1.493      \\
Fashion-MNIST & 240-260    & 0.726 & 1.502      & 1.467      \\ \bottomrule
\end{tabular}
\end{table}

 \section{Conclusion, limitations and future research} \label{sect:conclusion}
From our theoretical and experimental results, we exploit the data manifold to preserve the topology information of its perturbation. We implement the EMaP to realize the idea and demonstrates its benefits in the explaining task. We recognize the main limitation of EMaP is in its requirement of the low-dimensional representations of the data and the local affine subspaces. For more complex data, computing them correctly can be very challenging. There are several interesting open questions of EMaP that we leave for our future work. For instance, it is important to study the impact of the underlying manifold-learning algorithm, i.e. the UMAP, on the perturbations and the explanations. It is also interesting to examine the behavior of EMaP in a wider range of explainers and applications.

\vskip 0.2in
\bibliography{minh}

\clearpage

\appendix

\clearpage

\section{Visualizations of synthetic data and explanations generated with EMaP} \label{appendix:viz}

This Appendix provides some visualizations of our synthetic data (in experiments of Table~\ref{table:h0h1syn}) and explanations generated with or without EMaP. The explanations of EMaP shown in this Appendix are those used in the experiments on Section~\ref{sect:exp} of the main manuscripts.

 \begin{figure}[ht]
     \centering
         \includegraphics[width=0.99\linewidth]{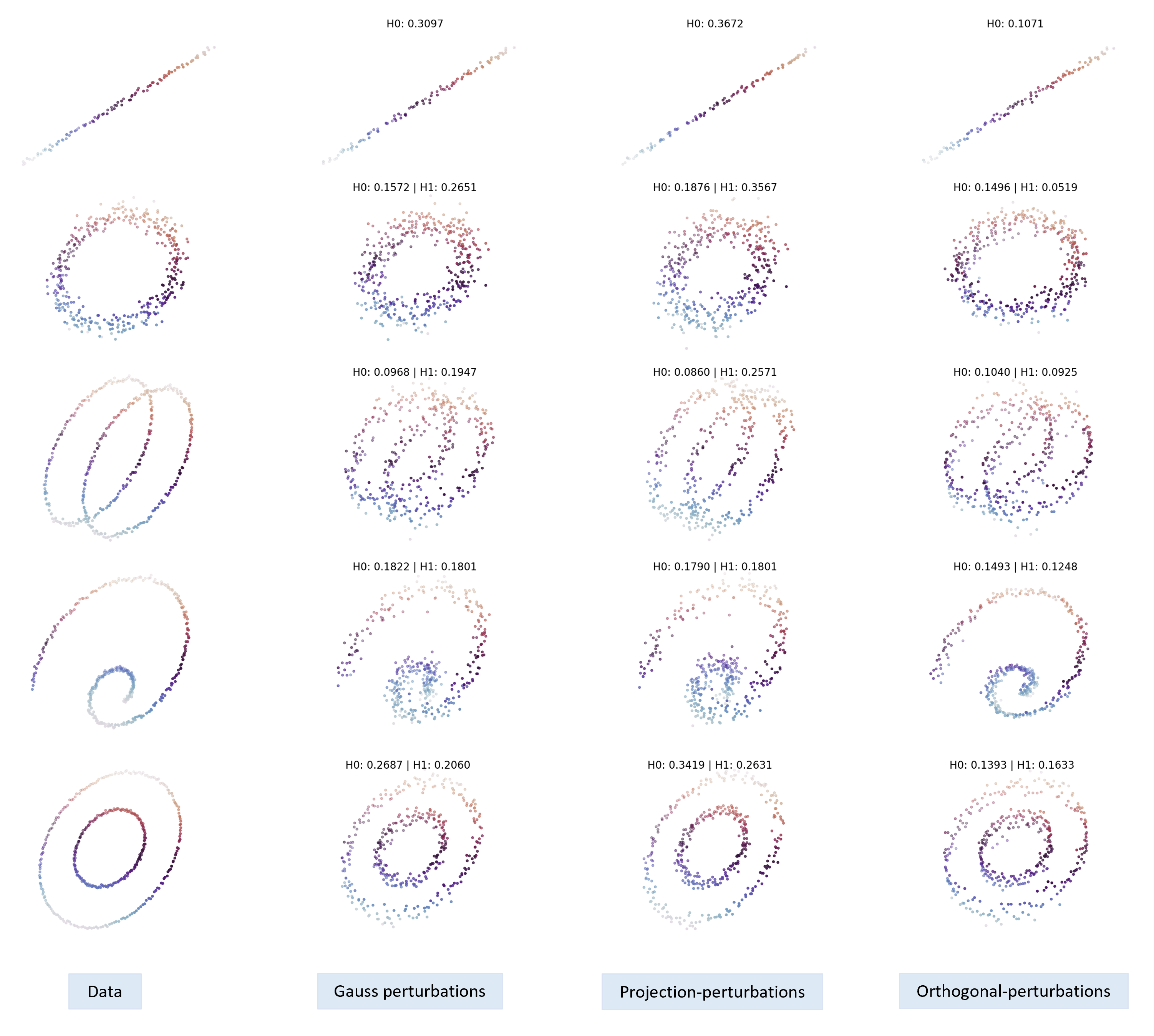}
     \caption{The visualization of some synthetic data in experiments of Table~\ref{table:h0h1syn}.}
     \label{fig:synthetic}
 \end{figure}

 In Fig.~\ref{fig:synthetic}, we visualize the synthetic data of different shapes and their perturbations in three dimensions. We also report $H_0$ and $H_1$ Bottleneck distances between the perturbations and the original data (left-column). 

 \clearpage

\begin{figure}[ht]
	\centering
	\begin{subfigure}{0.24\linewidth}
     \centering
         \includegraphics[height=0.92\linewidth]{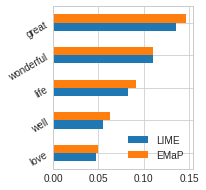}
         \caption*{Positive}
	\end{subfigure}
	\begin{subfigure}{0.24\linewidth}
     \centering
         \includegraphics[height=0.92\linewidth]{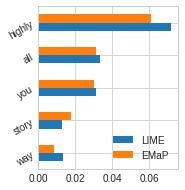}
         \caption*{Positive}
	\end{subfigure}
	\begin{subfigure}{0.24\linewidth}
     \centering
         \includegraphics[height=0.92\linewidth]{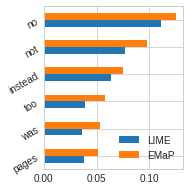}
         \caption*{Negative}
	\end{subfigure}
	\begin{subfigure}{0.24\linewidth}
     \centering
         \includegraphics[height=0.92\linewidth]{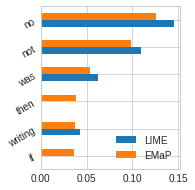}
         \caption*{Negative}
	\end{subfigure}
	\caption{Visualization of EMaP-LIME explanations of the Multi-polarity-books review datasets~\citep{multibook}.}
	\label{fig:viz_text}
\end{figure}

\begin{figure}[ht]
	\centering
	\begin{subfigure}{0.24\linewidth}
     \centering
         \includegraphics[height=0.92\linewidth]{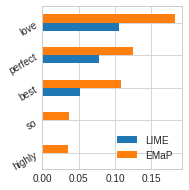}
         \caption*{Positive}
	\end{subfigure}
	\begin{subfigure}{0.24\linewidth}
     \centering
         \includegraphics[height=0.92\linewidth]{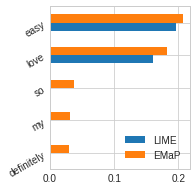}
         \caption*{Positive}
	\end{subfigure}
	\begin{subfigure}{0.24\linewidth}
     \centering
         \includegraphics[height=0.92\linewidth]{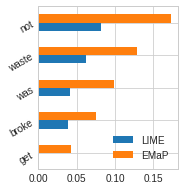}
         \caption*{Negative}
	\end{subfigure}
	\begin{subfigure}{0.24\linewidth}
     \centering
         \includegraphics[height=0.92\linewidth]{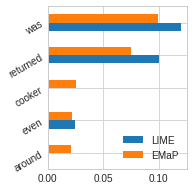}
         \caption*{Negative}
	\end{subfigure}
	\caption{Visualization of EMaP-LIME explanations of the Multi-polarity-kitchen review datasets~\citep{multibook}.}
	\label{fig:viz_kit}
\end{figure}

Fig.~\ref{fig:viz_text} and \ref{fig:viz_kit} compare the actual explanations returned by LIME and LIME with EMaP in the books' reviews in the Multi-Domain Sentiment datasets. We can see that the weights of features included in the explanations are quite similar between the two methods.

\begin{figure}[ht]
	\centering
	\begin{subfigure}{0.95\linewidth}
     \centering
         \includegraphics[width=1.0\linewidth]{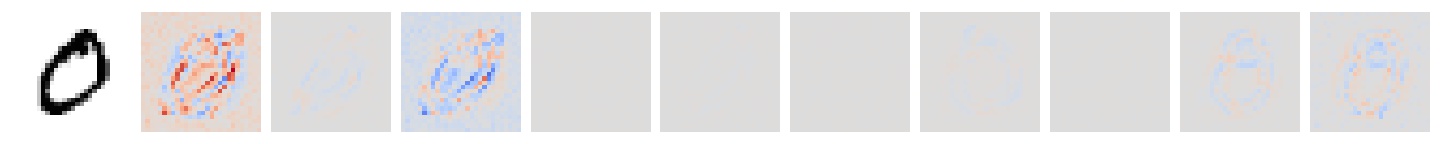}
	\end{subfigure}
	\begin{subfigure}{0.95\linewidth}
     \centering
         \includegraphics[width=1.0\linewidth]{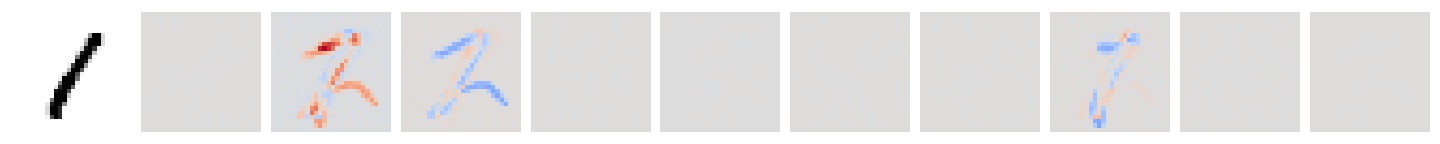}
	\end{subfigure}
	\begin{subfigure}{0.95\linewidth}
     \centering
         \includegraphics[width=1.0\linewidth]{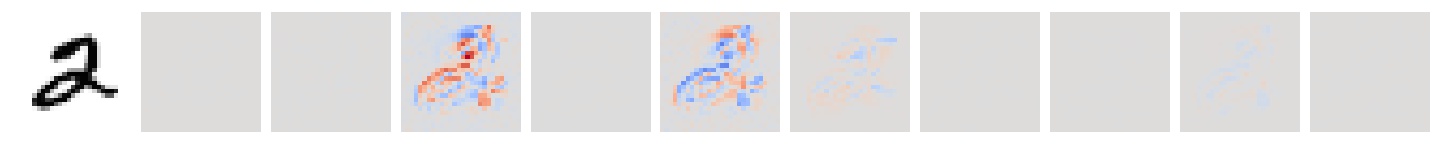}
	\end{subfigure}
	\begin{subfigure}{0.95\linewidth}
     \centering
         \includegraphics[width=1.0\linewidth]{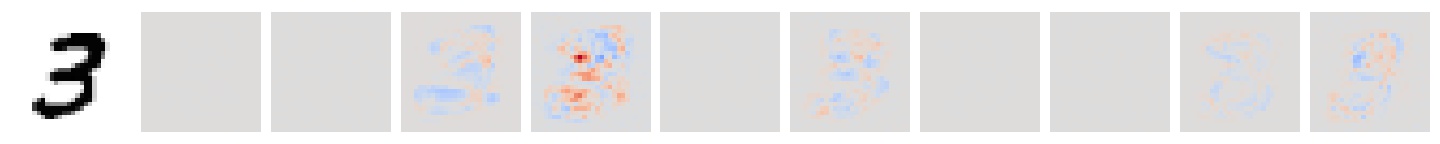}
	\end{subfigure}
	\begin{subfigure}{0.95\linewidth}
     \centering
         \includegraphics[width=1.0\linewidth]{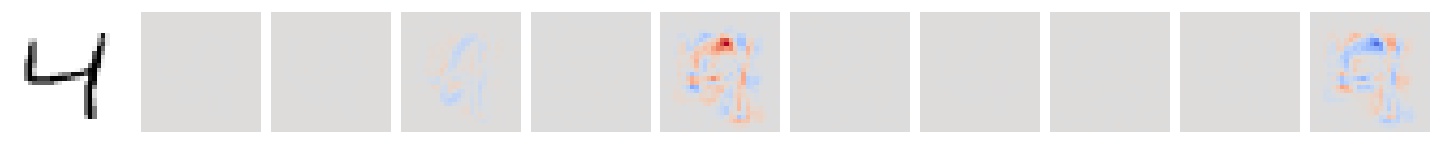}
	\end{subfigure}
	\begin{subfigure}{0.95\linewidth}
     \centering
         \includegraphics[width=1.0\linewidth]{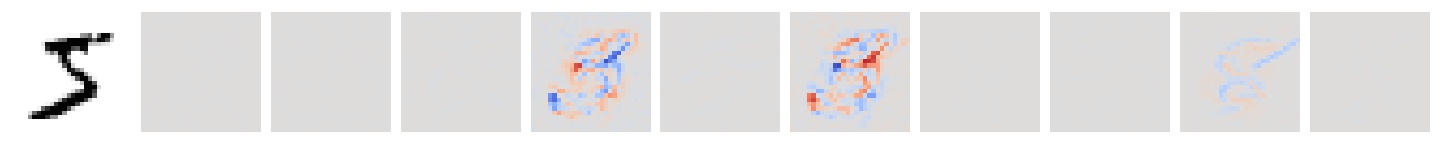}
	\end{subfigure}
	\begin{subfigure}{0.95\linewidth}
     \centering
         \includegraphics[width=1.0\linewidth]{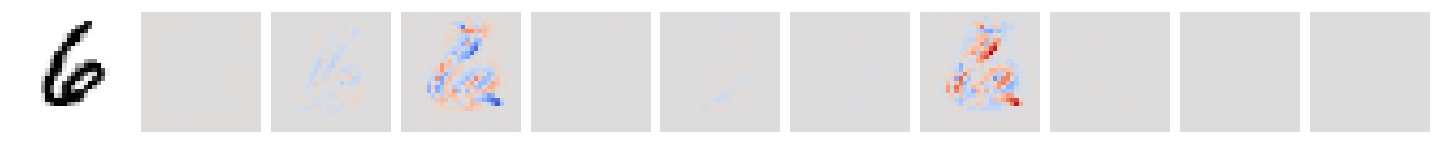}
	\end{subfigure}
	\begin{subfigure}{0.95\linewidth}
     \centering
         \includegraphics[width=1.0\linewidth]{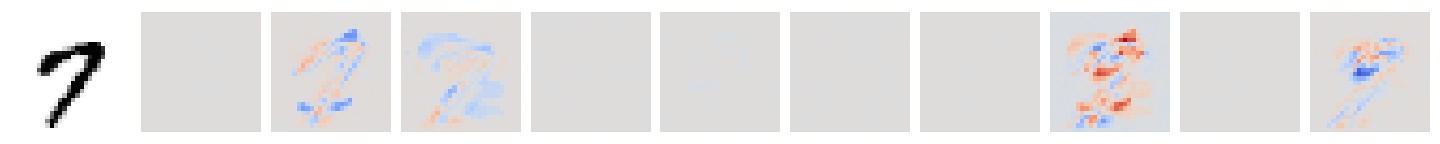}
	\end{subfigure}
	\begin{subfigure}{0.95\linewidth}
     \centering
         \includegraphics[width=1.0\linewidth]{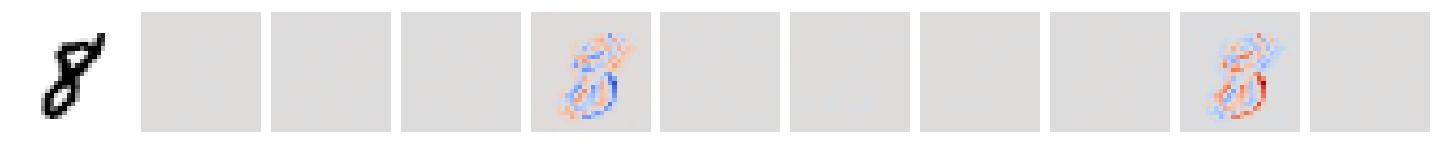}
	\end{subfigure}
	\begin{subfigure}{0.95\linewidth}
     \centering
         \includegraphics[width=1.0\linewidth]{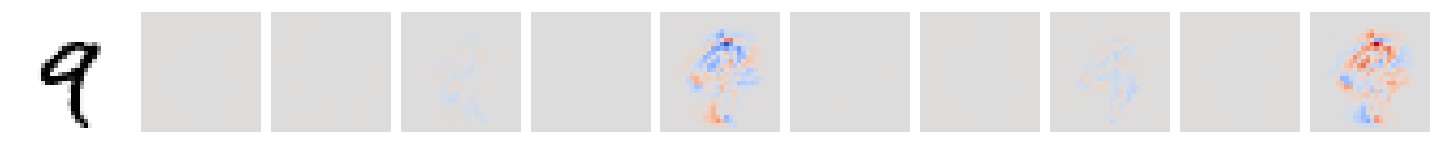}
	\end{subfigure}
	\caption{Visualization of EMaP-LIME explanations of the MNIST dataset. The column indicates the class label to be explained.}
	\label{fig:viz_mnist}
\end{figure}

Fig~\ref{fig:viz_mnist} shows the explanations of LIME with EMaP for all classes in the MNIST dataset. The red (blue) areas mean that, if the features in those areas remain unchanged (change), the activation of that class will be stronger.


\begin{figure}[ht]
	\centering
	\begin{subfigure}{0.95\linewidth}
     \centering
         \includegraphics[width=0.65\linewidth]{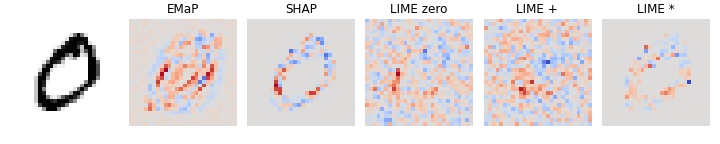}
	\end{subfigure}
	\begin{subfigure}{0.95\linewidth}
     \centering
         \includegraphics[width=0.65\linewidth]{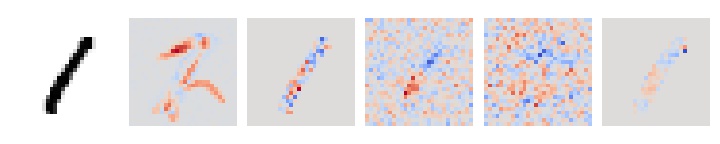}
	\end{subfigure}
	\begin{subfigure}{0.95\linewidth}
     \centering
         \includegraphics[width=0.65\linewidth]{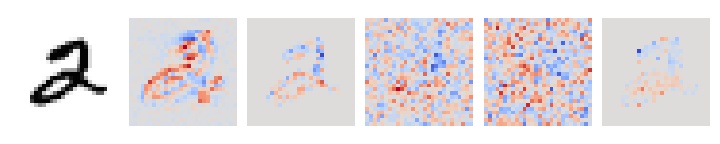}
	\end{subfigure}
	\begin{subfigure}{0.95\linewidth}
     \centering
         \includegraphics[width=0.65\linewidth]{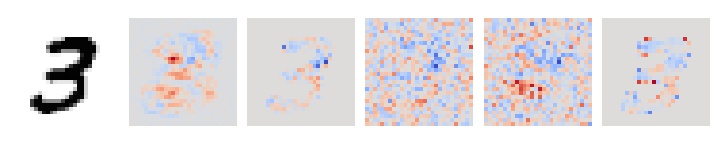}
	\end{subfigure}
	\begin{subfigure}{0.95\linewidth}
     \centering
         \includegraphics[width=0.65\linewidth]{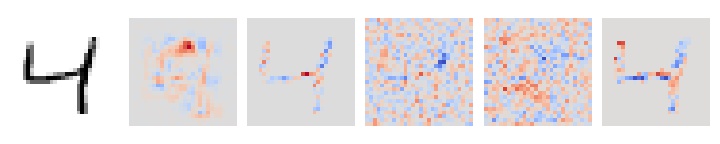}
	\end{subfigure}
	\begin{subfigure}{0.95\linewidth}
     \centering
         \includegraphics[width=0.65\linewidth]{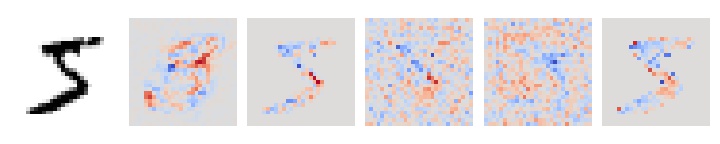}
	\end{subfigure}
	\begin{subfigure}{0.95\linewidth}
     \centering
         \includegraphics[width=0.65\linewidth]{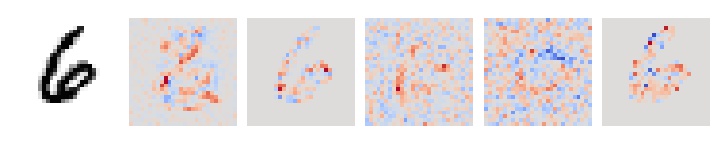}
	\end{subfigure}
	\begin{subfigure}{0.95\linewidth}
     \centering
         \includegraphics[width=0.65\linewidth]{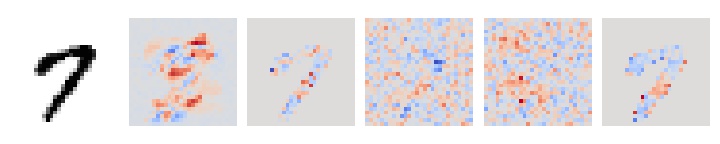}
	\end{subfigure}
	\begin{subfigure}{0.95\linewidth}
     \centering
         \includegraphics[width=0.65\linewidth]{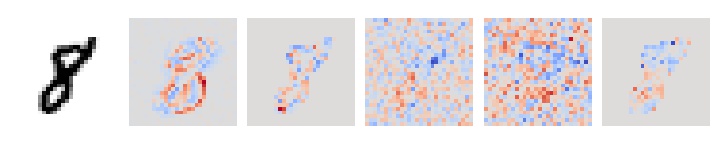}
	\end{subfigure}
	\begin{subfigure}{0.95\linewidth}
     \centering
         \includegraphics[width=0.65\linewidth]{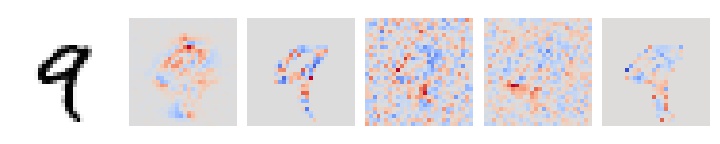}
	\end{subfigure}
	\caption{Examples of explanations in MNIST. Modifying the red-est (blue-est) area would negate (strengthen) the original prediction.}
	\label{fig:compare_examples_mnist}
\end{figure}

\begin{figure}[ht]
	\centering
	\begin{subfigure}{0.95\linewidth}
     \centering
         \includegraphics[width=0.65\linewidth]{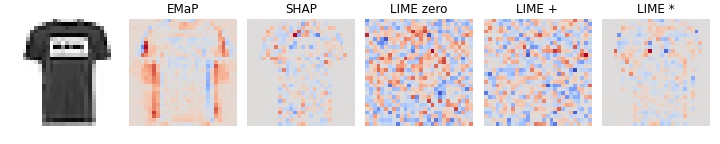}
	\end{subfigure}
	\begin{subfigure}{0.95\linewidth}
     \centering
         \includegraphics[width=0.65\linewidth]{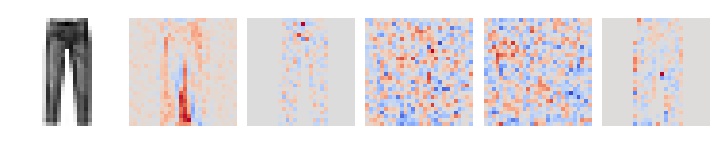}
	\end{subfigure}
	\begin{subfigure}{0.95\linewidth}
     \centering
         \includegraphics[width=0.65\linewidth]{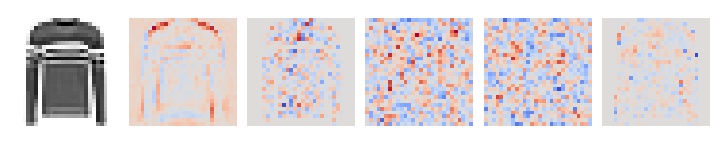}
	\end{subfigure}
	\begin{subfigure}{0.95\linewidth}
     \centering
         \includegraphics[width=0.65\linewidth]{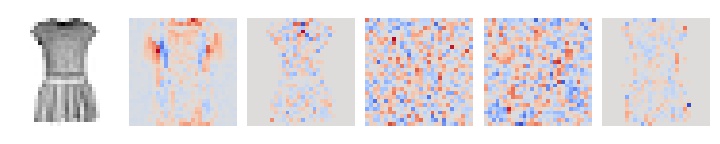}
	\end{subfigure}
	\begin{subfigure}{0.95\linewidth}
     \centering
         \includegraphics[width=0.65\linewidth]{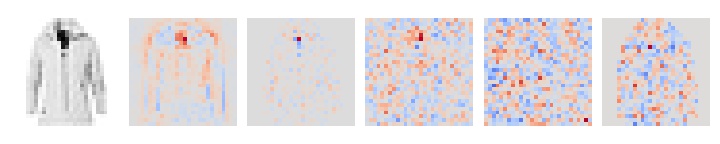}
	\end{subfigure}
	\begin{subfigure}{0.95\linewidth}
     \centering
         \includegraphics[width=0.65\linewidth]{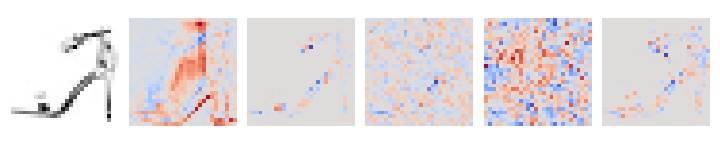}
	\end{subfigure}
	\begin{subfigure}{0.95\linewidth}
     \centering
         \includegraphics[width=0.65\linewidth]{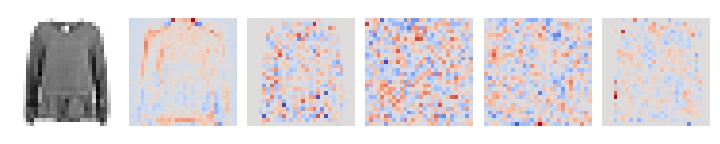}
	\end{subfigure}
	\begin{subfigure}{0.95\linewidth}
     \centering
         \includegraphics[width=0.65\linewidth]{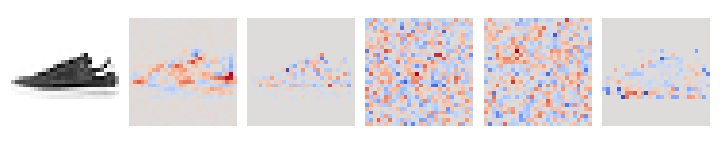}
	\end{subfigure}
	\begin{subfigure}{0.95\linewidth}
     \centering
         \includegraphics[width=0.65\linewidth]{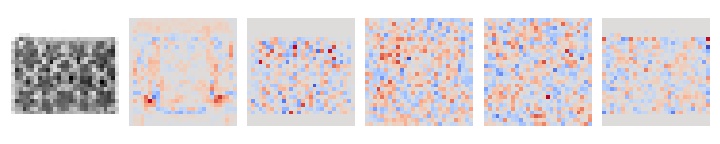}
	\end{subfigure}
	\begin{subfigure}{0.95\linewidth}
     \centering
         \includegraphics[width=0.65\linewidth]{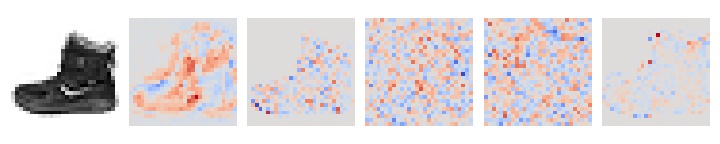}
	\end{subfigure}
	\caption{Examples of explanations in Fashion-MNIST. Modifying the red-est (blue-est) area would negate (strengthen) the original prediction.}
	\label{fig:compare_examples}
\end{figure}


 Fig.~\ref{fig:compare_examples_mnist} and \ref{fig:compare_examples} provide some explanations of EMaP along with explanations of other methods in MNIST and Fashion-MNIST,respectively.

\end{document}